\documentclass[a4paper,fleqn]{cas-dc}

\usepackage[authoryear]{natbib}
\usepackage{hyperref}
\usepackage{graphicx}
\usepackage{subfigure}
\usepackage{epstopdf}
\usepackage{color}
\usepackage[noend]{algpseudocode}
\usepackage{algorithmicx,algorithm}

\usepackage{amsmath,amsfonts}
\usepackage{array}
\usepackage{textcomp}
\usepackage{stfloats}
\usepackage{url}
\usepackage{verbatim}
\usepackage{cite}
\usepackage{booktabs}
\usepackage{multirow}
\usepackage{threeparttable}
\usepackage[normalem]{ulem}
\useunder{\uline}{\ul}{}
\usepackage{float} 
\usepackage{utfsym}

\usepackage{hyperref}
\usepackage[nameinlink]{cleveref}
\def\tsc#1{\csdef{#1}{\textsc{\lowercase{#1}}\xspace}}
\tsc{WGM}
\tsc{QE}
\tsc{EP}
\tsc{PMS}
\tsc{BEC}
\tsc{DE}
\begin{document}
\begin{sloppypar}

	\let\WriteBookmarks\relax
	\def\floatpagepagefraction{1}
	\def\textpagefraction{.001}
	\let\printorcid\relax
	\shorttitle{}
	\shortauthors{X.Y. Zhang et~al.} %% 缩略作者 自己名字， 比如： 张三 = S. Zhang

	%% 标题
	\title [mode = title]{Learning states enhanced knowledge tracing: Simulating the diversity in real-world learning process}
    % 学习状态增强的知识追踪:模拟真实世界学习过程中的差异性
	
 %%\tnotemark[1,2]
	%%\tnotetext[1]{This document is the results of the research project funded by the National Science Foundation.}

	%%\tnotetext[2]{The second title footnote which is a longer text matter to fill through the whole text width and overflow into another line in the footnotes area of the first page.}

	% \cormark[1]%%通讯作者星标
	% \fnmark[1] %%第几作者
	%\credit{}%%本文的贡献
	%% 作者顺序
	\author[1]{\textcolor[RGB]{0,0,1}{Shanshan Wang}}
	\ead{wang.shanshan@ahu.edu.cn}
	\author[1]{\textcolor[RGB]{0,0,1}{Xueying Zhang}}
	\ead{q22301200@stu.ahu.edu.cn}
	\author[2]{\textcolor[RGB]{0,0,1}{Xun Yang}}
    \cormark[1]%%通讯作者星标
	\ead{xyang21@ustc.edu.cn}
    \author[3]{\textcolor[RGB]{0,0,1}{Xingyi Zhang}}
	\ead{xyzhanghust@gmail.com}
    \author[4]{\textcolor[RGB]{0,0,1}{Keyang Wang}}
	\ead{20181202010t@alu.cqu.edu.cn}

 \address[1]{Information Materials and Intelligent Sensing Laboratory of Anhui Province, Anhui University, Anhui 230601, China}
 \address[2]{School of Data Science, University of Science and Technology of China, Anhui 230026, China}
 \address[3]{College of Computer Science, Sichuan University, Chengdu 610065, China}
 \address[4]{Zhejiang Dahua Technology Co., Ltd, Hangzhou 310053, China}

	\cortext[cor1]{Corresponding author.} %% 首页左下角通讯作者
	%%\cortext[cor2]{Principal corresponding author} 

	% \fntext[fn1]{Equal Contribution.}
	%%\fntext[fn2]{Another author footnote, this is a very long footnote and it should be a really long footnote. But this footnote is not yet sufficiently long enough to make two lines of footnote text.}

	%%\nonumnote{This note has no numbers. In this work we demonstrate $a_b$ the formation Y\_1 of a new type of polariton on the interface between a cuprous oxide slab and a polystyrene micro-sphere placed on the slab.}

    % \small
	%%摘要
\begin{abstract}
 The Knowledge Tracing (KT) task focuses on predicting a learner's future performance based on the historical interactions. The knowledge state plays a key role in learning process. However, considering that the knowledge state is influenced by various learning factors in the interaction process, such as the exercises similarities, responses reliability and the learner's learning state. Previous models still face two major limitations. First, due to the exercises differences caused by various complex reasons and the unreliability of responses caused by guessing behavior, it is hard to locate the historical interaction which is most relevant to the current answered exercise. Second, the learning state is also a key factor to influence the knowledge state, which is always ignored by previous methods. To address these issues, we propose a new method named Learning State Enhanced Knowledge Tracing (LSKT). Firstly, to simulate the potential differences in interactions, inspired by Item Response Theory~(IRT) paradigm, we designed three different embedding methods ranging from coarse-grained to fine-grained views and conduct comparative analysis on them. Secondly, we design a learning state extraction module to capture the changing learning state during the learning process of the learner. In turn, with the help of the extracted learning state, a more detailed knowledge state could be captured. Experimental results on four real-world datasets show that our LSKT method outperforms the current state-of-the-art methods.
 
% we could develop a learning state enhanced knowledge state extraction module, in this way,
 \end{abstract}
   % 知识追踪(KT)任务侧重于基于历史交互预测学习者未来的表现。知识状态在学习过程中起着关键作用。然而，考虑到知识状态在交互过程中受到各种学习因素的影响，如练习相似度、反应信度和学习者的学习状态。以前的模型仍然面临两个主要限制。首先，由于各种复杂原因导致的练习差异和猜测行为导致的回答不可靠，很难找到与当前回答的练习最相关的历史交互作用。其次，学习状态也是影响知识状态的关键因素，这一点在以往的方法中往往被忽略。为了解决这些问题，我们提出了一种新的方法——学习状态增强知识跟踪(LSKT)。首先，为了模拟交互中的潜在差异，受项目反应理论(IRT)范式的启发，我们设计了从粗粒度视图到细粒度视图的三种不同嵌入方法，并对它们进行了对比分析。其次，我们设计了一个学习状态提取模块来捕捉学习者在学习过程中学习状态的变化。反过来，在提取的学习状态的帮助下，可以捕获更详细的知识状态。在四个真实数据集上的实验结果表明，我们的LSKT方法优于当前最先进的方法。

	% 	%%\includegraphics{./grabs.pdf} %%图片摘要地址路径

	%%高亮
	% \begin{highlights}
	% 	\item End-to-end community detection method based on graph convolution network.
	% 	\item A new community perspective similarity is proposed.
	% 	\item Modify the convolution layer for large networks.
	% 	\item The loss function based on modularity and Bernoulli Poisson model is introduced.
	% 	\item Evaluate performance using real-world networks.
	% \end{highlights}
		
	%% 关键词
% Keywords: 智慧教育,知识追踪,教育数据挖掘,深度学习。

	% 此指令为生成标题格式，不可删除
	\maketitle

	%% 1.引言
	\section{Introduction}
	%%\par{文本内容}换行并缩进
\par{
%With the rapid development of online educational services, Knowledge Tracing (KT)~\citep{[28]} has become a crucial means to enhance teaching effectiveness. By analyzing learners' learning data, KT can predict their future performance, enabling educators to more accurately meet learners' learning needs and provide personalized learning resources, thus improving teaching efficiency. Driven by large-scaled online open courses~(MOOCs) and other educational platforms~\citep{[26]}, KT technology has experienced rapid development, providing robust support for intelligent online education. 
%It effectively addresses the shortcomings of learner feedback and teaching resources, making educational services better suited to the practical needs of learners and teachers~\citep{[29],[30],[31]}
}
% 随着在线教育服务的飞速发展，知识跟踪（KT）[28]已经成为提升教学效果的重要手段。通过分析学习者的学习数据，KT能够预测学习者的未来表现，使教育工作者能够更准确地满足学习者的学习需求，提供个性化的学习资源,从而提高教学效率。在大规模在线开放课程（MOOCs）等教育平台的推动下[26][27]，KT技术得到了快速发展，为智能在线教育提供了强大的支持。它有效地弥补了学习者反馈和教学资源的不足，使教育服务更能满足学习者和教师的实际需求[29][30][31]。
\par{
Knowledge Tracing (KT) is a challenging task as the real learning process of humans involves numerous complex learning behaviors and is influenced by various factors, including the learning state during answering, the difficulty of exercises, and tendencies towards guessing~\citep{[32]} and so on. The key to KT task lies in comprehensively simulating these complex factors and effectively modeling them as real-world learning process.
}
% 知识追踪（Knowledge Tracing，KT）是一项极具挑战性的任务，因为人类的真实学习过程涉及众多复杂的学习行为，并受到各种因素的影响，包括答题时的学习状态、练习的难度、猜测倾向~citep{[32]}等。KT 任务的关键在于全面模拟这些复杂因素，并将其有效地建模为真实世界的学习过程。

\par{ 
In recent years, the Transformer has shown great potential in the field of KT ~\citep{[42],[43]}. Many Deep Learning-based Knowledge Tracing (DLKT) models, such as Attention-based DLKT (ATT-DLKT) models, adopts the Transformer to capture the inherent relationships between learners' historical interactions, to accurately estimate their knowledge states. However, these models often have some limitations. Firstly, they usually rely too much on learners' historical performance on similar exercises to assess their knowledge states~\citep{[2],[13],[14]}. Moreover, to alleviate data sparsity problem, many models choose the Knowledge Concepts (KCs) instead of exercises for model training, thus the rich association information between exercises and interactions could be lost. This inevitably increases the difficulty for ATT-DLKT models to accurately identify the key historical moments and may introduce noise into the model's training. Additionally, due to  subjectivity, the unreliable responses caused by learner guessing factors could also inevitably bring in some noise in the interaction. Secondly, the learner's changing learning state is another important factor in the learning process which is always ignored by previous methods.
This states is related with the learner's recent performance and can complement the knowledge state, together influencing the learner's next performance. 
To address these issues, we propose a new ATT-DLKT method named Learning State Enhanced Knowledge Tracing (LSKT). This approach, on one hand, further mines the potential interaction information from exercises and responses to obtain the more precise feature embedding. On the other hand, incorporates the changes in learners' answering process into the capturing of knowledge states. Thereby the performance of our model could be improved.
}
% 近年来，变换器（Transformer）在知识追踪（KT）领域展现出了巨大的潜力~\citep{[42], [43]}。许多基于深度学习的知识追踪（DLKT）模型，如基于注意力的 DLKT（ATT-DLKT）模型，都采用变换器来捕捉学习者历史交互之间的内在关系，从而准确估计学习者的知识状态。然而，这些模型往往存在一些局限性。首先，它们通常过于依赖学习者在类似练习中的历史表现来评估他们的知识状态~/citep{[2],[13],[14]} 。此外，为了缓解数据稀疏的问题，许多模型选择知识概念（KC）而不是习题来训练模型，因此习题与交互之间丰富的关联信息可能会丢失。这不可避免地增加了 ATT-DLKT 模型准确识别关键历史时刻的难度，并可能在模型训练中引入噪声。此外，由于主观性，学习者猜测因素导致的不可靠反应也不可避免地会在交互中带来一些噪声。其次，学习者不断变化的学习状态是学习过程中的另一个重要因素，而以往的方法总是忽略这一点。这种状态与学习者最近的表现相关，可以补充知识状态，共同影响学习者的下一次表现。为了解决这些问题，我们提出了一种新的 ATT-DLKT 方法，即学习状态增强知识追踪法。

% \vspace{1.85em}
% \hspace{3.7em}

\begin{figure*}
  \begin{minipage}[b]{0.535\textwidth}
    \centering
    \includegraphics[width=0.965\linewidth]{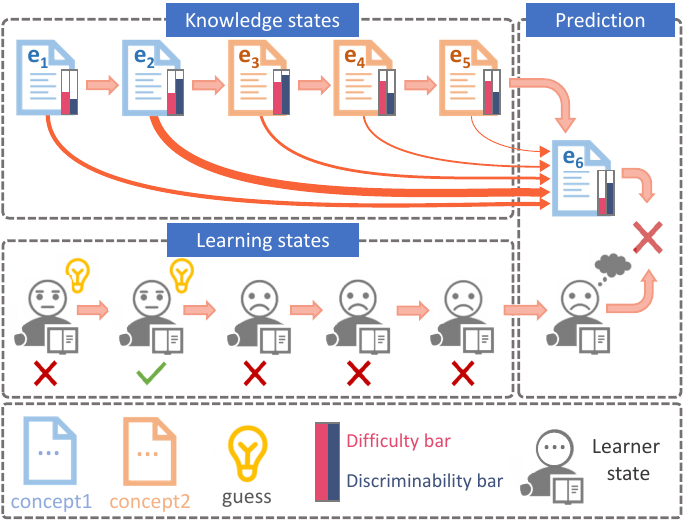}
    \raisebox{-0.3em}{(a) Motivation diagram}
  \end{minipage}
  \begin{minipage}[b]{0.45\textwidth}
    \centering
    \includegraphics[width=1.02\linewidth]{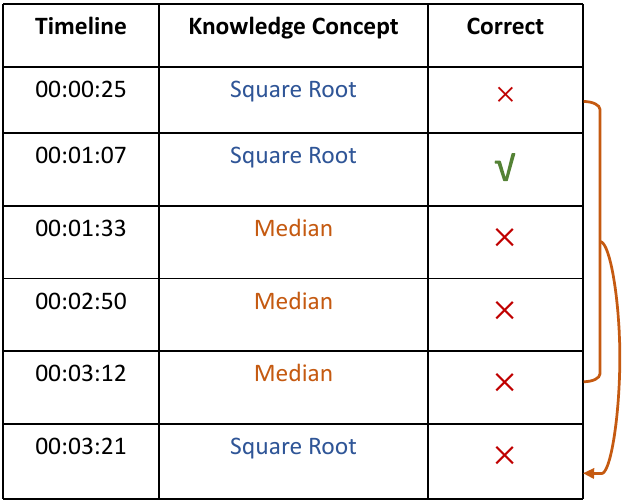}
    \raisebox{-0.3em}{(b) A learning segment in ASSIST12}
  \end{minipage}
\caption{Figure (a) illustrates the learning process of a learner. As shown in the figure, during the process of answering these six exercises, the learner is influenced by many factors. We need to observe the learner's performance in $e_{1}$ to $e_{5}$, infer their latent knowledge state and learning state, and consider both aspects to predict the learner's performance in exercise $e_{6}$. Figure (b) shows a real slice of the ASSIST12 dataset. It demonstrates that recent learning states have an impact on subsequent performance, even though different concepts are involved, which should be taken into account.
% 图(a)展示了一位学习者的学习过程。如图所示，在回答这六个练习的过程中，学习者受到许多因素的影响。我们需要观察学习者在$e_{1}$至$e_{5}$中的表现，推断他们的潜在知识状态和学习状态，并结合这两方面考虑来预测学习者在练习$e_{6}$中的表现。图（b）展示了 ASSIST12数据集上的一个真实切片。显示最近的学习状态对后续表现有影响，尽管涉及不同的概念，但应该予以考虑。
}
\label{FIG:1}
\vspace{-1.2em}
\end{figure*}

\par{
To better illustrate the above points, we provide a simple example in \Cref{FIG:1}. \Cref{FIG:1}(a) depicts a scenario where a learner answering six consecutive exercises, revealing two main factors influencing learner performance: knowledge state and learning state. Although exercises $e_{1}$ and $e_{2}$ involve the same knowledge concepts, the various factors including differences difficulty in exercise, discriminability in exercises and the learner's learning state all could affect the performance. From the traditional perspective of knowledge state, exercise $e_{2}$, which shares more similarities in influencing factors with $e_{6}$, would provide more valuable predictions for $e_{6}$; indicating that $e_{6}$ is more likely to be predicted as correct. However, even if $e_{2}$ is answered correctly, there still could be possibilities of lucky guesses. Therefore, only relying the similarity may the prediction on $e_{6}$ be wrong. Further more, the learning state is also a crucial factor which could influence the learner performance. From the perspective of learning state evolution, poor performance on exercises $e_{1}$ - $e_{5}$ may lead to a decline in the learner's learning state, which could affect their performance on $e_{6}$. In fact, the real answer in $e_{6}$ is incorrectly, which implies that during model training, we need to fully consider the knowledge states and learning states to make the model closer to the real-world answering process. \Cref{FIG:1}(b) shows a real slice from the ASSIST12 dataset, detailing the information of $e_{1}$-$e_{6}$ as depicted in the left panel. From this slice, it's evident that even with different adjacent interaction knowledge concepts, a learner's recent performance could influence their future performance. This observation is also supported by data analysis in literature~\citep{[1]}. However, past studies often overlooked the complex factors during the answering process. This prompts us to consider how to effectively model these factors to fully utilize the model's potential, capturing more fine-grained changes in knowledge states, and thus more realistically reflecting the learner's answering process.
}
%为了更好地说明以上几点，我们在\Cref{FIG:1}中提供了一个简单的示例。\Cref{FIG:1} (a)描述了一个学习者回答六个连续练习的场景，揭示了影响学习者表现的两个主要因素:知识状态和学习状态。虽然练习$e_{1}$和$e_{2}$涉及相同的知识概念，但练习难度的不同、练习的可辨别性、学习者的学习状态等各种因素都会影响成绩。从传统的知识状态角度来看，运动$e_{2}$与$e_{6}$在影响因素上有更多的相似之处，对$e_{6}$的预测更有价值;表明$e_{6}$更有可能被预测为正确的。然而，即使$e_{2}$的答案是正确的，仍然有可能是幸运的猜测。因此，仅依靠相似性，$e_{6}$上的预测可能是错误的。此外，学习状态也是影响学习者学习成绩的重要因素。从学习状态演化的角度来看，在练习$e_{1}$ - $e_{5}$上表现不佳可能导致学习者的学习状态下降，从而影响他们在$e_{6}$上的表现。事实上，$e_{6}$中的真实答案是不正确的，这意味着在模型训练时，我们需要充分考虑知识状态和学习状态，使模型更接近现实世界的回答过程。\Cref{FIG:1} (b)显示了ASSIST12数据集的真实切片，详细说明了左侧面板中所示的$e_{1}$ - $e_{6}$的信息。从这张图中可以明显看出，即使相邻的交互知识概念不同，学习者最近的表现也会影响他们未来的表现。这一观察结果也得到了文献\citep{[1]}数据分析的支持。然而，以往的研究往往忽略了回答过程中的复杂因素。这促使我们考虑如何有效地对这些因素进行建模，以充分利用模型的潜力，捕捉更细粒度的知识状态变化，从而更真实地反映学习者的回答过程。

\par{To address the issues mentioned above, we proposed LSKT based on ATT-DLKT model. There are three main contributions in this method. Firstly,three feature embedding methods are designed from coarse-grained to fine-grained inspired by IRT paradigm. By combining the potential differences in interactions, the model could not only mitigate the overfitting problem, but also capture the refined  difference between interactions. Additionally, we also explored the impact of different embedding paradigm on the performance of the model. Secondly, to extract the changing learning state during the learning process, dues to the ability of natural capturing capability of the interaction information within each moment, the causal convolution layers with different receptive field sizes are leveraged. In this way, the short-term changes and possible patterns in the learning process can be captured, which were overlooked by previous ATT-DLKT models. 
Thirdly, considering that the learning state is a key fact to influence the knowledge states, we aim to incorporate the learning state into the process of extracting knowledge states, and then a learning state-enhanced knowledge state extraction module is designed. 

In summary, our LSKT method could integrate global contextual information, capture long-range dependencies, and introduce sparse attention to the learning state. In this way, the detailed knowledge states and learning states are captured jointly without introducing additional noise and the key contributions are as follows:
}
% 为了解决上述问题，我们提出了基于at - dlkt模型的LSKT。这种方法有三个主要贡献。首先，受IRT范式的启发，设计了从粗粒度到细粒度的三种特征嵌入方法;通过结合交互中的个性化信息，该模型不仅可以缓解过拟合问题，而且可以捕捉到交互之间的精细差异。此外，我们还探讨了不同嵌入范式对模型性能的影响。
% 其次，为了提取学习过程中不断变化的学习状态，利用每一时刻交互信息的自然捕获能力，利用不同感受野大小的因果卷积层;通过这种方式，可以捕获学习过程中的短期变化和可能的模式，这是以前的at - dlkt模型所忽略的。
% 第三，考虑到学习状态是影响知识状态的关键因素，将学习状态引入到知识状态提取过程中，设计了学习状态增强的知识状态提取模块。
% 总之，我们的LSKT方法可以整合全局上下文信息，捕获远程依赖关系，并引入对学习状态的稀疏关注。这样，在不引入额外噪声的情况下，可以同时捕获详细的知识状态和学习状态，主要贡献如下:

\par{The key contributions can be summarized as follows:}

\begin{enumerate}[(1)]
\item{We reveal a potential issue in ATT-DLKT model, which relies heavily on learners' past performances of similar exercises to assess their knowledge state, while failing to adequately consider the dynamic changes in learning state. By integrating the consideration of both states, the model's predictive accuracy and applicability can be significantly improved, making it more consistent with actual learning environments.}
% 1）我们揭示了 ATT-DLKT 模型的一个潜在问题，即该模型严重依赖学习者过去在类似练习中的表现来评估其知识状态，而未能充分考虑学习状态的动态变化。通过综合考虑这两种状态，可以显著提高模型的预测准确性和适用性，使其更符合实际学习环境。
\item{Inspired by IRT, three different personalized level on interaction embedding is designed and compared, simulating realistic differences between interactions. This design not only alleviates the issue of model overfitting caused by data sparsity but also enhances the model's performance and interpretability.}
% 2）受IRT启发，设计并对比了三种不同粒度的习题与交互嵌入方式，模拟了练习之间与交互之间的真实差异。这种设计不仅缓解了数据稀疏性导致的模型过拟合问题，同时提升了模型效果和可解释性。

\item{We proposed a novel learning state extraction module which always ignored by previous methods. This module could capture the learner's learning patterns over multiple time scales during the answering process, so it can effectively represent the learner's learning state at history answer time.
}
% 我们提出了一个新颖的学习状态提取模块，这是以往方法一直忽略的。该模块可以捕捉学习者在答题过程中多个时间尺度上的学习模式，因此能有效地代表学习者在历史答题时间的学习状态。

\item{To capture the precise knowledge state, a learning state-enhanced knowledge state extraction module has been proposed. This module achieves more fine-grained knowledge state by paying sparse attention to the learning states in the process of tracing knowledge states.
}
% 为了精确捕捉知识状态，我们提出了一个学习状态增强型知识状态提取模块。该模块在追踪知识状态的过程中，通过对学习状态的稀疏关注，实现了更精细的知识状态。

\end{enumerate}
\par{
The subsequent sections of this article are organized as follows: The second part reviews and analyzes related work. The third part defines KT problems and introduces the specific approach of our model (LSKT). The fourth part reports the experimental results on four public datasets and discusses and summarizes our method.
% 本文的后续部分组织如下：第二部分对相关工作进行了回顾和分析。第三部分定义了KT问题，并介绍了我们模型（LSKT）的具体方法。第四部分报告了在四个公共数据集上的实验结果，并对本研究进行了讨论与总结。
}
    \section{Related work}
	\label{Related_work}
% 	\par{
% 		In this section, we will analyze and review representative works in the field from two perspectives: classical KT methods and attention-based KT methods.
% % 在本节中，我们将从经典的KT方法与基于注意力的KT方法两方面来分析和回顾该领域的现有代表性工作。
% 	}
 \subsection{Traditional knowledge  tracing}
	\par{
 Probabilistic models and logical models were two categories of early knowledge tracing models. Probabilistic models used Markov processes (HMM)~\citep{[4]}. The Bayesian Knowledge Tracing (BKT) method was a classic example of this~\citep{[5]}. This method was proposed by Corbett and Anderson. In this method, the mastery state of knowledge points was modeled as a binary variable (mastered/not mastered), and the learning process was modeled as discrete transitions from the not mastered state to the mastered state. With further research on BKT, subsequent studies incorporated more influencing factors, such as temporal differences in data~\citep{[8]}, hierarchical relationships between knowledge points~\citep{[9]}, exercise difficulty~\citep{[10]}, etc. logical models were based on logical functions and used continuous distributions instead of discrete probabilities to represent learners' knowledge states, thus better capturing learning intensity. The basic principle involved calculating the probability of correctly answering exercises based on learner learning ability parameters and exercise parameters (such as difficulty and discrimination). Performance Factors Analysis (PFA)~\citep{[6]} and Learning Factors Analysis (LFA)~\citep{[7]} were two classic logical models for knowledge tracing. While these traditional knowledge tracing models possess strong interpretability, they often exhibit certain limitations and biases as they rely on domain knowledge annotated by experts as input features. Such models tend to be somewhat one-sided and constrained, making it challenging to fully uncover the hidden information in the data. Consequently, the predictive performance of early models is typically subpar.
% 早期的知识追踪模型可以分为概率模型和逻辑模型两类。概率模型基于马尔可夫过程(HMM)[4]，其中最经典的是由Corbett和Anderson提出的贝叶斯知识追踪（BKT）方法。在该方法中，知识点的掌握状态被建模为二进制变量（掌握/未掌握），并将学习过程建模为从未掌握状态到掌握状态的离散转换。随着对BKT的研究深入，后续的研究将更多的影响因素纳入考虑，如数据的时间差异信息[8]、知识点之间的层次关系[9]、习题的难度[10]等。逻辑模型基于逻辑函数，使用连续分布而不是离散的概率来表示学习者的知识状态，因此能够能更好地表示学习强度。其基本原理是通过学习者学习能力参数和练习参数（如难度、区分度）来计算正确回答练习的概率。绩效因素分析（PFA）[6]和学习因素分析模型（LFA）[7]是两种经典的知识追踪逻辑模型。虽然这些传统的知识追踪模型具有较强的可解释性，但由于需要依赖专家标注的领域知识作为输入特征，这种人为构建的模型往往具有一定的片面性和局限性，难以充分挖掘数据中隐藏的信息，因此，早期模型的预测效果往往一般。
	}
 \par{
Deep learning is getting a lot of attention from researchers because of its excellent ability to extract features. This has led to a new field called Deep Learning Knowledge Tracing (DLKT). Currently, DLKT models mainly adopt three network architectures: Recurrent Neural Networks (RNNs)~\citep{[41]}, Memory Networks~\citep{[16]}, and Attention Networks~\citep{[15]}. Deep Knowledge Tracing (DKT)~\citep{[11]} was among the first models to base knowledge tracing on deep learning, with one common strategy involving the use of Recurrent Neural Networks. In the DKT model, the hidden units of the RNN were utilized to represent the learner's knowledge state, which was updated as the learner's answering behavior evolved. Another approach was the Dynamic Key-Value Memory Networks (DKVMN)~\citep{[12]}, which employed memory networks to depict the learner's knowledge state. This involved using static "key" matrices to represent fixed knowledge concepts and dynamic "value" matrices to reflect the progression of knowledge states. However, the efficacy of DKT and DKVMN was found to be inadequate when dealing with sparse historical data and limited interactions between learners and exercises. To counter this challenge, Pandey et al. introduced a knowledge tracing framework based on self-attention mechanisms, termed Self-Attentive Knowledge Tracing (SAKT)~\citep{[13]}. This method was capable of recognizing the state of specific knowledge based on learners' past interactions with exercises and making predictions accordingly.
% 深度学习因其出色的特征提取能力逐渐成为了研究者们的关注重点,从而产生了基于深度学习的知识追踪（DLKT）领域。目前，DLKT模型主要采用了三种网络结构，分别是递归神经网络[41]、记忆网络[16]以及注意力网络[15]。深度知识追踪（DKT）[11]是第一个基于深度学习的知识追踪模型，它的一种典型方法是利用递归神经网络。在DKT模型中，递归神经网络的隐藏单元被用来代表学习者的知识状态，随着学习者答题情况的变化，这个知识状态表征也将会被更新。另一种典型的方法是动态键值记忆网络（DKVMN）[12]，它使用记忆网络来表征学习者的知识状态，静态的“键”矩阵和动态的“值”矩阵被用来表示固定的知识概念和知识状态的发展。然而，当面临历史数据稀疏、学习者与习题交互有限的情况时，DKT和DKVMN的性能表现并不理想。为了解决这个问题，Pandey等人提出了一种基于自我注意力机制的知识追踪框架，即自我注意知识追踪（SAKT）[13]，该方法能够通过学习者的历史习题交互情况，识别出特定知识的状态，并据此进行预测。
 }
 \subsection{Transformer-based knowledge tracing}
 \par{
 In recent years, the outstanding performance of Transformer models in fields such as NLP and multimodal information processing~\citep{[33],[34],[35]} has sparked researchers' interest in their application to the field of KT.  Early studies, such as~\citep{[13]}, primarily employed the self-attention mechanism to capture learners' historical learning in order to infer their knowledge state. However, due to significant differences between KT datasets and natural language data, these models did not surpass the performance of traditional DLKT models, such as DKT and DKVMN. Recent research has started to tackle this issue, with models like AKT~\citep{[2]}, DTransformer~\citep{[14]}, and FKT~\citep{[37]}. AKT improved model performance by incorporating a monotonic attention mechanism to model learners' forgetting behavior and using embeddings from the Rasch model to capture differences among problems. DTransformer introduced contrastive learning to maintain the stability of knowledge states, alleviating the information bias issues observed in earlier studies. FKT adopted an encoder-decoder-predictor framework and integrated speed prediction as an additional task, enabling fine-grained knowledge tracing. These latest studies have focused on the characteristics of learner-item interaction data, enhancing Transformer-based KT methods, not only improving model performance but also advancing the application of Transformer models in the KT domain.
 % 近年来，Transformer模型在自然语言处理和多模态信息处理等领域的出色表现[33][34][35]，引发了研究者们对于其在知识追踪（KT）领域应用的兴趣。早期的研究，如文献[13]，主要利用自注意力机制来捕捉学习者的学习历史，从而推测其知识状态。但由于KT的数据集与自然语言数据存在较大差异，这些模型的性能并未超越传统的DLKT模型，如DKT和DKVMN。然而，最近的研究已经开始解决这个问题，例如AKT[2]、DTransformer[14]和FKT[37]等。AKT通过设计一种单调注意力机制来模拟学习者的遗忘行为，并使用Rasch模型的嵌入来捕捉题目间的差异性，从而提升模型性能。DTransformer引入了对比学习来保持知识状态的稳定性，减轻了早期研究中信息偏差的问题。FKT则采用了一个编码器-解码器-预测器框架，并将速度预测作为附加任务，实现了细粒度的知识追踪。这些最新的研究关注学习者与习题的交互数据特点，对基于Transformer的知识追踪方法进行了改进，不仅提升了模型性能，也推动了Transformer模型在知识追踪领域的应用。
 }
 \par{
 However, despite efforts by studies like AKT and FGKT~\citep{[3]} to consider the actual differences between exercises, limitations persist. For instance, AKT had only distinguished exercises with the same concept by simulating exercise difficulty, but considering only exercise difficulty information could not fully simulate the subtle differences between exercises, thereby limiting the model's development potential. FGKT uses three traditional fusion functions to model the differences between exercises and concepts, but its lack of theoretical basis makes the model's interpretability poor. To address these issues, our LSKT designs three different granularity differential exercise and interaction embeddings based on item response theory as inputs to the model, and conducted experimental comparisons, thereby improving the performance and interpretability of the model. Additionally, these models had often overlooked an important factor in the learning process, namely, the changes in learners' learning states. Our LSKT introduces consideration of learning states into the process of tracing learners' knowledge states, and combines the two states to jointly predict learners' future exercise performance.
% 然而，尽管像AKT和FGKT[3]这样的研究也尝试过考虑练习之间的实际差异，但仍有局限性。例如，AKT仅通过模拟习题难度来区分具有相同概念的习题，但是仅考虑习题难度信息无法充分模拟习题之间的细微差异，从而限制了模型的发展潜力。而FGKT使用三种传统的融合函数来建模习题与概念间的差异性，但其缺乏理论依据，使得模型的可解释性较差。为了解决上述问题，LSKT设计了基于项目反应理论的三种不同粒度的差异性练习嵌入作为模型的输入，并且进行了实验比较，从而提升了模型的性能与可解释性。此外，这些模型往往忽略了学习过程中一个重要的影响因素，即学习者学习状态的变化，LSKT将对学习状态的考虑引入学习者知识状态的追踪过程，并结合两种状态共同参与预测学习者未来的做题表现。
}
 \section{METHODOLOGY}
 \par{
This section will introduce our LSKT model. Firstly, the KT task will be clearly defined. Then, we'll provide an overview of LSKT's architecture and detail the three different feature embedding methods we've designed. Next, the learning state extraction module will be introduced and the knowledge state extraction module with enhanced learning states will be discussed in depth. Finally, the final prediction results will be generated through a prediction layer.

% 本节将深入探讨LSKT模型。首先，明确定义KT任务。接着，我们将简要概述LSKT的总体架构，并详细说明我们设计的三种不同粒度的特征嵌入方法。然后，介绍学习状态提取模块并深入讨论学习状态增强的知识状态提取模块。最后，将通过一个预测层生成最终的预测结果。
 }
\subsection{Problem definition}
\par{
In this section, we present the definition of the KT task and summarize the main parameters' symbols utilized throughout the paper in \Cref{Table:1}.
% 在本节中，我们给出了KT任务的定义，并在表1中总结了整篇论文中使用的主要符号。
In practical learning scenarios, learners typically answer exercises in the recommended order of the educational system. This interaction process can be represented as $K=\left\{k_{1}, k_{2}, \ldots, k_{n}\right\}$, where $n \in \mathbb{N}^{+}$ represents the number of exercises. For a specific learner, the interaction with exercises can be represented as a triplet $k_{t} =\left(e_{t}, c_{t}, r_{t}\right)$, where $e_{t} \in \mathbb{N}^{+}$ and $c_{t} \in \mathbb{N}^{+}$ denote the exercise index and concept index at time $t$, respectively, and $r_{t} \in\{0,1\}$ represents the response at time $t$ (0 for incorrect and 1 for correct). Since we primarily focus on predictions for individual learners, for readability, we omit the learner index. Thus, the answering process of each learner can be represented as the following sequence:
}
\begin{equation}
\label{deqn_ex1a}
K=\left\{\left(e_{1}, c_{1}, r_{1}\right), \ldots,\left(e_{t}, c_{t}, r_{t}\right)\right\} 
\end{equation}
\par{
The goals of our LSKT can be primarily divided into two parts. One is to obtain the more accurate knowledge state. Specifically, we propose the learning state $\left\{\hat{y}_{1}, \hat{y}_{2}, \ldots, \hat{y}_{t}\right\}$ up to time $t$ based on the interaction sequence $K$ of the learner.  By fusing with the original knowledge state $\left\{h_{1}, h_{2}, \ldots, h_{t}\right\}$, the fused state $\left\{z_{1}, z_{2}, \ldots, z_{t}\right\}$ is obtained. The other is to predict the learner's performance on the task at time $t+1$, denoted as $\hat{r}_{t+1}$. Our LSKT utilize the fused state $z_{t}$ above and exercise $e_{t}$ are leveraged to participate the downstream task.
}
% 在实际的学习场景中，学习者通常会按照教育系统的推荐顺序回答问题。这个交互过程可以被表示为$K=\left\{k_{1}, k_{2}, \ldots, k_{n}\right\}$，其中$n \in \mathbb{N}^{+}$代表题目数量。对于特定的学习者，他们与题目的交互可以被表示为一个三元组$k_{t} =\left(e_{t}, c_{t}, r_{t}\right)$，$e_{t} \in \mathbb{N}^{+}$和$c_{t} \in \mathbb{N}^{+}$分别代表在t时刻的练习索引和概念索引，而$r_{t} \in\{0,1\}$表示在t时刻的回答（0代表错误，1代表正确）。由于我们主要关注对单个学习者的预测，为了便于阅读，我们省略了学习者索引。因此，每个学习者的答题过程可以被表示为如下序列：
% $K=\left\{\left(e_{1}, c_{1}, r_{1}\right), \ldots,\left(e_{t}, c_{t}, r_{t}\right)\right\} \quad e_{t} \in \mathbb{N}^{+}, c_{t} \in \mathbb{N}^{+}, r_{t} \in\{0,1\}$
% LSKT的目标主要分为两部分。首先，它需要根据学习者的交互序列K来跟踪t时刻及其之前的知识状态$\left\{h_{1}, h_{2}, \ldots, h_{t}\right\}$和学习状态$\left\{\hat{y}_{1}, \hat{y}_{2}, \ldots, \hat{y}_{t}\right\}$。然后，将这两种状态融合，得到融合状态$\left\{z_{1}, z_{2}, \ldots, z_{t}\right\}$。其次，LSKT会利用融合状态$z_{t}$和习题$e_{t}$来预测$t+1$时刻学习者的答题表现$\hat{r}_{t+1}$。

\begin{table}[]
\centering
\caption{Notations and explanations.}
\label{Table:1}
\footnotesize
\begin{tabular}{ll}
\hline
Notations                                                & Explanations                                                       \\ \hline
$e$                                                      & Exercise index                                                     \\
$c$                                                      & Concept index                                            \\
$r$                                                      & Response index                                                     \\
$c_{c}$,$c_{c}^{\prime}$                                 & Concept embedding and its variation                                \\
$\alpha_{e}$                                             & Deviation degree parameter                                         \\
$r_{r}$,$r_{r}^{\prime}$                                 & Response embedding and its variation                               \\
$g_{\left(c, r\right)}$,$g_{\left(c, r\right)}^{\prime}$ & Original interaction embedding and its variation                   \\
$d_{e}$                                                  & Discrimination feature                                             \\
$f_{c}$                                                  & Guessing factor feature                                           \\
$x$                                                      & Exercise feature                                                   \\
$y$                                                      & Interaction feature                                                \\
$\hat{y}$                                                & Learning state feature                                              \\
$\beta$                                              & Learning state-based similarity distribution  \\
$\gamma_{t, \tau}$                                       & Complete attention scores at time $t$ versus time $\tau$                    \\
$h$                                                      & Knowledge state                                                    \\
$z$                                                      & Synthesis of knowledge state and learning state                      \\
$\hat{r}_{t+1}$                                          & Prediction at time $t+1$                                           \\ \hline
\end{tabular}
\end{table}

\begin{figure*}
\centering
{\includegraphics[width=6.2in]{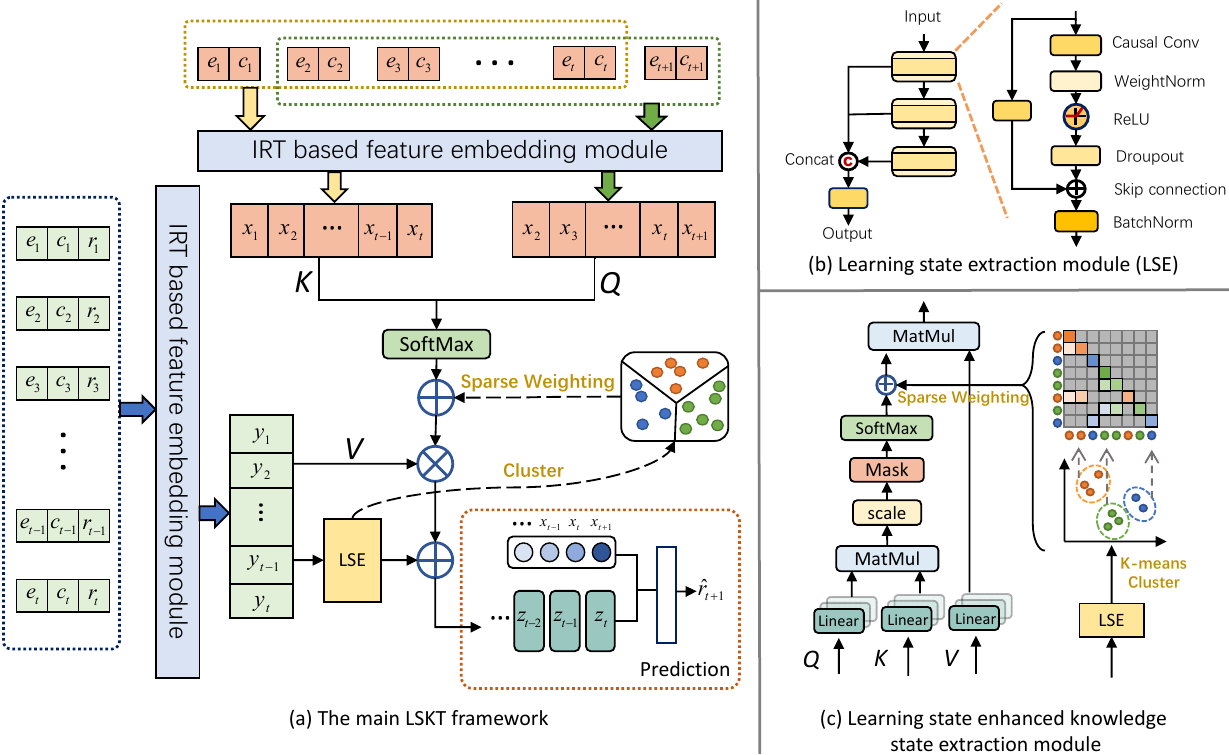}}
\vspace{1em}
\caption{(a) The overall architecture of our LSKT. (b) The learning state extraction module, where we extract and integrate multi-scale learning state information through multi-layer causal convolutions. (c) The learning state enhanced knowledge state extraction module}
% 图1：(a)LSKT的整体架构。(b)学习状态提取模块，我们通过多层的因果卷积提取并融合多尺度的学习状态信息。(c)学习状态增强的知识状态提取模块。
\label{FIG:2}
\vspace{-1.3em}
\end{figure*}

\subsection{Model overview}
\par{
The diagram in \Cref{FIG:2} illustrates the backbone network of our LSKT along with its components. Our LSKT model consists of four parts: the IRT based feature embedding module, the Learning State Extraction (LSE) module, the learning state enhanced knowledge state extraction module, and the Learner Response Prediction module. Firstly, through the feature embedding module, embedded representations of learners' exercise and interaction features are obtained. Then, the LSE module is utilized to extract the sequence of learning state changes from the embedded interaction sequences of learners. Subsequently, we employ sparse attention calculation based on the k-means clustering method for each moment in the sequence of learning state changes. This allows us to obtain sparse attention scores by masking irrelevant moments of learning state and integrating them into the process of capturing knowledge states, thereby emphasizing similar moments of learning state. Finally, the learning state and knowledge state are fused together to jointly predict learners' performance in answering exercises in the next moment. Unlike the previous ATT-DLKT method, our model simulates the potential differences in interactions and pays attention to the changes in the learner's learning state during the answering process. By introducing consideration of these two key points into the model, it more finely simulates the process of knowledge acquisition in the real world, thereby enhancing the model's effectiveness and interpretability.
% 图二展示了LSKT的主干网络与其内部的各个组件。LSKT模型由四个部分组成，分别是基于IRT的特征融合模块，学习状态提取模块（LSE），学习状态增强的知识状态提取模块与学习者回答预测模块。首先，通过特征融合模块，我们获取了学习者的习题和交互特征的嵌入表示。然后，利用LSE模块从学习者的交互嵌入序列中提取学习状态变化序列。接着，我们对学习状态变化序列中的每个时刻进行基于k-means聚类方法的稀疏注意力计算，通过屏蔽学习状态不相关时刻获得对于学习状态的稀疏注意力关注，并将其融入到知识状态的捕获过程中，实现对相似学习状态时刻的强调。最后，将学习状态和知识状态融合，共同参与预测学习者下一时刻的答题表现。与过去的ATT-DLKT方法不同，我们的模型模拟了交互间的潜在差异信息并关注了学习者答题过程中学习状态的变化，通过在模型中引入对上述两个要点的考虑，更细致地模拟了真实世界中的知识获取过程，从而提高了模型的效果和可解释性。
}
\subsection{IRT based feature embedding module}
\par{
In real world educational environment, the number of questions in question banks often far exceeds the number of learners, leading to many questions being answered by only a few learners, resulting in data sparsity issues. To address this problem, many models utilize knowledge concepts to index questions, thus avoiding overfitting. However, these methods often overlook the potential differences embedded in learner interactions, which include the potential distinctions between different exercises under the same knowledge concept, as well as the unreliability of responses caused by learners' guessing behavior. This oversight undoubtedly limits the potential of knowledge tracing (KT) methods.
% 在现实的教育环境中，题库的问题数量常常远超学习者的数量，导致许多问题仅被少数学习者回答，引起数据稀疏性问题。为解决此问题，许多模型采用知识概念对问题进行索引，以避免过度拟合。然而，这些方法往往忽视了学习者交互中蕴含的潜在差异，这些差异既包括同一知识概念下不同习题之间的潜在区别，也包括学习者猜测行为导致的回答的不确定性。这种忽视无疑限制了知识追踪（KT）方法的潜能。
}
\par{
To address the above issues, methods for the sequence of exercises and interactions with learners are delved into. We believe that the three-parameter model of Item Response Theory (IRT) can progressively uncover the impact of differences between exercises and interactions on learner performance. Inspired by this, exercise embeddings and interaction embeddings corresponding to these three different levels of refinement in parameter modeling are designed. Our design effectively alleviates the problem of model overfitting and gradually explores the subtle differences between different exercises and interactions under the same concept, thereby fully unleashing the potential of the model.The effects of the three modeling methods on model training in RQ4 will be further compared.
% 为了解决上述问题，深入的探究了习题序列与学习者交互序列的建模方法。我们认为IRT的三个参数模型能够渐进的挖掘出交互之间的潜在差异对于学习者表现的影响。受此启发，设计了对应于这三种不同精细程度参数模型的习题嵌入与交互嵌入方法，我们的设计有效减轻了模型的过拟合问题，并逐步探索了同一概念下不同习题与交互之间的微小差异，从而充分发挥模型的潜力。将在RQ4中进一步比较三种建模方法对模型训练的影响。
}
\par{
  \textbf{1PL-based embeddings.} The LSKT-1PL model corresponds to the IRT one-parameter model, which introduces a difficulty difference parameter among exercises to achieve coarse-grained modeling of exercise characteristics and interaction features. Specifically, the modeling of exercise characteristics is as follows:
% 基于1PL的嵌入：LSKT-1PL模型对应于IRT一参数模型，通过引入习题间的困难度差异参数，我们实现了粗粒度的习题特征与交互特征建模。其中，习题特征建模如下：
{\small
\begin{equation}
\label{deqn_ex1a}
x_{t}=\left[c_{c_{t}} \| \left( \alpha_{e_{t}} \cdot c_{c_{t}}^{\prime}\right) \right]W_{1}
\end{equation}
}
where $x_{t} \in \mathbb{R}^{D}$ represents the feature vector of the exercise at time $t$. $c_{c_{t}}, c_{c_{t}}^{'} \in \mathbb{R}^{D}$ denote the $D$-dimensional continuous vectors obtained for the knowledge concept $c_{t}$ through a Knowledge concept feature extractor, and the corresponding variation obtained through another Knowledge concept feature extractor, respectively. $\alpha_{e_{t}} \in \mathbb{R}$ is a learnable scalar parameter representing the difficulty of the exercise $e_{t}$, and we use $\alpha_{e_{t}} \cdot c_{c_{t}}^{\prime}$ to simulate the difficulty differences among different exercises. The symbol $\|$ denotes the feature concatenation operation, and $W_{1} \in \mathbb{R}^{2D\times D}$ is a learnable parameter matrix. For simplicity of the formula expression, we omit the bias parameter required for the dimensionality reduction operation.
% 在这个公式中，$x_{t} \in \mathbb{R}^{D}$ 表示$t$时刻的的习题特征$c_{c_{t}},c_{c_{t}}^{'} \in \mathbb{R}^{\mathrm{D}}$  分别代表知识概念 $c_{t}$通过知识概念特征提取器得到的$D$维连续向量与c_{t}通过另一知识概念特征提取器得到的对应变化量。$\alpha_{e_{t}} \in \mathbb{R}$ 是根据练习 $e_{t}$得到的一个可学习难度标量参数，我们利用 $\alpha_{e_{t}} \cdot c_{c_{t}}^{\prime}$ 来模拟不同练习之间的困难度差异。$\|$表示特征拼接操作，$W_{1} \in \mathbb{R}^{2D\times D}$是一个可学习的参数矩阵，为了简化公式的表达，我们省略了执行降维操作所需的偏移量参数。
}
\par{
Similarly, the coarse-grained modeling of learner-exercise interaction features is as follows:
% 相似的，粗粒度的学习者与习题的交互特征建模如下：
{\small
\begin{align}
\small
&g_{\left(c_{t}, r_{t}\right)} = c_{c_{t}} + r_{r_{t}} \\
&g_{\left(c_{t}, r_{t}\right)}^{\prime} = c_{c_{t}}^{\prime} + r_{r_{t}}^{\prime}\\
&y_{t}=\left[g_{\left(c_{t}, r_{t}\right)} \| \left( \alpha_{e_{t}} \cdot g_{\left(c_{t}, r_{t}\right)}^{\prime} \right) \right]W_{2}
\end{align}
}
where $r_{r_{t}}, r_{r_{t}}^{\prime} \in \mathbb{R}^{\mathrm{D}}$ represent the $D$-dimensional continuous vectors obtained from the learner's response $r_{t}$ through two response different feature extractors, with $r_{r_{t}}$ denoting the original embedding and $r_{r_{t}}^{\prime}$ denoting the corresponding change. $g_{\left(c_{t}, r_{t}\right)}, g_{\left(c_{t}, r_{t}\right)}^{\prime} \in \mathbb{R}^{\mathrm{D}}$ represent the original interaction embedding between context $c_{t}$ and response $r_{t}$, and their corresponding changes, respectively. $W_{2} \in \mathbb{R}^{2D\times D}$ is a learnable parameter matrix. $y_{t} \in \mathbb{R}^{\mathrm{D}}$ represents the interaction feature between the learner and the exercise at time step t.
% 在上述公式中，$r_{r_{t}},r_{r_{t}}^{\prime} \in \mathbb{R}^{\mathrm{D}}$ 分别代表从学习者的回答r_{t}通过响应特征提取器得到的$D$维连续嵌入向量与r_{t}通过另一响应特征提取器得到的对应变化量。$g_{\left(c_{t}, r_{t}\right)},g_{\left(c_{t}, r_{t}\right)}^{\prime} \in \mathbb{R}^{\mathrm{D}}$ 分别表示原始的学习者交互嵌入与其对应变化量。$W_{2} \in \mathbb{R}^{2D\times D}$是一个可学习的参数矩阵。$y_{t} \in \mathbb{R}^{\mathrm{D}}$ ，表示做题过程中在t时刻学习者与习题的交互特征。
}
\par{
\textbf{2PL-based embeddings.} LSKT-2PL corresponds to the IRT two-parameter model, which introduces a discrimination parameter between exercises on the basis of LSKT-1PL, achieving a sub-fine-grained modeling of the differences between exercises and 
between interactions.
% 基于2PL的嵌入：LSKT-2PL对应于IRT二参数模型，其在LSKT-1PL的基础上引入了习题间区分度参数，实现了次细粒度的习题之间与交互之间的差异性建模：
{\small
\begin{align}
&x_{t}\mkern-3mu=c_{c_{t}}\mkern-3mu +\mkern-2mu \left[\operatorname{Repeat}\left(\mkern-3mu\alpha_{e_{t}}, D\mkern-3mu\right) \mkern-3mu\| \left( W_{3}\mkern-3mu \cdot d_{e_{t}} \right) \right]W_{4} \mkern-3mu\cdot c_{c_{t}}^{\prime} \\
&y_{t}\mkern-3mu=g_{\left(c_{t}, r_{t}\right)}\mkern-3mu + \mkern-3mu\left[\operatorname{Repeat}\left(\mkern-3mu\alpha_{e_{t}}, D\mkern-3mu\right) \mkern-3mu\| \left( W_{3} \mkern-4mu\cdot d_{e_{t}} \right) \right]W_{5} \mkern-2mu \cdot \mkern-3mu g_{\left(c_{t}, r_{t}\right)}^{\prime}
\end{align}
}
where $\operatorname{Repeat}\left(\cdot, D\right)$ represents the repetition of the difficulty scalar parameter to obtain a D-dimensional vector. $d_{e_{t}} \in \mathbb{R}^{D}$ denotes a D-dimensional mapping of exercises in the latent space. $W_{3} \in \mathbb{R}^{D\times D}$ is a learnable parameter matrix. We use $ W_{3} \cdot d_{e_{t}}$ to represent a finer differentiation between exercises, namely the distinctiveness parameter of exercises. Building upon LSKT-1PL, LSKT-2PL integrates the effects of two parameters, namely the difficulty and distinctiveness, on the embedding of exercises at time $t$ and the interaction embedding. $W_{4}, W_{5} \in \mathbb{R}^{2D\times D}$ represent two learnable dimension reduction parameter matrices.
% 在上述公式中，$\operatorname{Repeat}\left(\cdot, D\right)$代表重复难度标量参数得到D维向量。$d_{e_{t}} \in \mathbb{R}^{D}$ 表示习题在隐藏空间中的一个 D 维映射,$W_{3} \in \mathbb{R}^{D\times D}$是一个可学习的参数矩阵，我们用$ W_{3} \cdot d_{e_{t}}$代表习题间更细粒度的差异性区分，即习题的区分度参数。在LSKT-1PL的基础上LSKT-2PL融合了对习题难度与区分度两个参数对 t 时刻习题嵌入以及交互嵌入的影响。$W_{4}, W_{5} \in \mathbb{R}^{2D\times D}$代表两个可学习的降维参数矩阵。
}
\par{
\textbf{3PL-based embeddings.} LSKT-3PL corresponds to the IRT three-parameter model. It builds upon LSKT-2PL by introducing the possibility of learners guessing, which is often overlooked when modeling the answering process. In reality, the impact of answering a exercise truthfully versus guessing should differ in terms of the learner's knowledge state. To simulate learners' guessing behavior, we incorporated random guessing perturbations into the learner interaction sequences, achieving a fine-grained model embedding:
% 基于3PL的嵌入：LSKT-3PL对应于IRT三参数模型，其在LSKT-2PL的基础上引入了学习者猜测的可能性，这也是在建模的答题过程时经常被忽略掉的一点。实际上，真实回答一道练习和猜测一道练习对于学习者知识状态的影响应该是不同的。为了模拟学习者的猜测行为，我们在学习者交互序列中加入了随机的猜测扰动，实现了细粒度的模型嵌入建模：
{\small
\begin{align}
&f_{c_{t+1}}=c_{c_{t+1}}+\operatorname{Random}\left\{0, \tilde{r}_{\text {Random }\{0,1\}}\right\} \\
&y_{t}\mkern-3mu=\mkern-3mu f_{c_{t+1}}\mkern-9mu + \mkern-3mu g_{\left(c_{t}, r_{t}\right)}\mkern-4mu + \mkern-3mu\left[\operatorname{Repeat}\left(\mkern-3mu\alpha_{e_{t}}, D\mkern-3mu\right) \mkern-3mu\| \mkern-3mu \left( W_{3}\mkern-4mu\cdot d_{e_{t}} \mkern-3mu \right) \right] \mkern-3mu W_{6} \mkern-3mu \cdot \mkern-3mu g_{\left(c_{t}, r_{t}\right)}^{\prime}
\end{align}}
where $\operatorname{Random}\left\{0, \tilde{r}_{\text {Random }\{0,1\}}\right\}$ represents randomly selecting whether to introduce a guessing factor, where 0 indicates no guessing, and $\tilde{r}_{\text {Random }\{0,1\}} \in \mathbb{R}^{D}$ represents randomly introducing a D-dimensional learner response embedding vector. Here, a 0-valued embedding vector represents an incorrect guess, while a 1-valued embedding vector represents a correct guess. In the entire formula, $f_{c_{t+1}}$ denotes whether to introduce the guessing factor for the next exercise, indicating the possibility of guessing or not guessing, and the possibility of guessing correctly or incorrectly. We choose to model the guessing factor for the next exercise because the goal of Knowledge Tracing (KT) is to predict the learner's performance on the next exercise based on the current knowledge state. Therefore, the knowledge state at time $t$ should include the guessing factor for time $t+1$. $W_{6} \in \mathbb{R}^{2D\times D}$ is a learnable parameter matrix. The learner's actual answering process can be better simulated by introducing the guessing factor in the interaction sequence $y_{t}$. The 3PL embedding modeling does not modify the exercise feature $x_{t}$ based on the 2PL embedding modeling.
% 在上述公式中，$\operatorname{Random}\left\{0, \tilde{r}_{\text {Random }\{0,1\}}\right\}$ 表示随机选择是否引入猜测因素，其中0代表不进行猜测，而$\tilde{r}_{\text {Random }\{0,1\}} \in \mathbb{R}^{D}$ 则表示随机引入一个D维学习者回答嵌入向量猜测，即0值嵌入向量代表猜测错误，1值嵌入向量代表猜测正确。在整个公式中，$f_{c_{t+1}}$ 的含义是对于下一题是否引入猜测因素，即可能进行猜测也可能不进行猜测，且猜测结果可能正确也可能错误。我们选择建模下一题的猜测因素是因为知识追踪（KT）的目标是通过当前的知识状态来预测学习者在下一题的表现。因此，时刻$t$的知识状态应该包含对于时刻$t+1$的猜测因素。$W_{6} \in \mathbb{R}^{2D\times D}$是一个可学习的参数矩阵。通过在交互序列$y_{t}$ 中引入猜测因素，我们可以更好地模拟学习者真实的答题过程。3PL嵌入建模不对习题特征x_{t}在2PL嵌入建模的基础上进行改动。
}
\subsection{Learning state extraction module (LSE)}
\par{
In the actual answering process, the learner's state is changing, influenced by various complex factors. \textit{e.g.}, continuous wrong answers might dent their confidence, making them more prone to errors even when facing questions they have not fully mastered. A study~\citep{[1]} suggests that learners' recent performance significantly impacts their next steps in real test environments. However, existing ATT-DLKT models often overlook this aspect. Therefore, to capture learners' learning states, we've devised the LSE module.
% 在实际的答题过程中，学习者的状态是变化的，这种变化受到许多复杂因素的影响。例如，如果学习者连续答错题目，可能会打击他们的自信心，导致在面对没有完全掌握的题目时更容易犯错误。一项研究[1]表明，在实际答题环境中，学习者的近期表现是影响他们下一步表现的关键因素。然而，现有的ATT-DLKT模型往往没有考虑到这一点。因此，为了捕捉到学习者的学习状态，我们设计了LSE模块。
}
\par{
\Cref{FIG:2}(b)  illustrates the structure of LSE. The structure of LSE consists of three residual blocks arranged sequentially to capture the learner's learning states at different scales, and finally achieves feature fusion through skip connections. The $1 \times 1$ convolutional layer is utilized for dimensionality reduction, yielding the comprehensive learning state of the learner. Each residual block comprises a causal convolutional layer, weight normalization layer, ReLU function, dropout layer, and skip connection. Layer normalization is applied between adjacent residual blocks.

The primary function of LSE is to perform causal convolution operations on the learner's historical interaction sequences. Since causal convolution strictly relies on past temporal information for prediction, for a historical interaction sequence $y$ and a one-dimensional convolutional kernel $s$, the mathematical expression of the causal convolution process can be represented as:
% 图2.(b)展示了LSE的结构。LSE由三个残差块串行组成，分别捕获学习者不同尺度的即近期学习状态，并最终通过跳跃连接完成特征融合。最后通过一个$1 \times 1$的卷积层降维，得到答题者的综合学习状态。每个残差块包含因果卷积层、权重归一化层、Relu函数、丢弃层和跳跃连接。相邻残差块之间采用层归一化处理。

% LSE的主要功能是对学习者历史交互序列进行因果卷积操作，由于因果卷积严格依赖于当前时刻之前的时序信息进行预测，对于历史交互序列$y$，以及一维卷积核$s$，因果卷积过程的的数学表达式可以表示为：
{\small
\begin{equation}
\tilde{y}_{t}=\sum_{m=0}^{M} y_{t-m} \cdot s_{M-m}
\label{Eq:10}
\end{equation}
}
\par{
Here, $\hat{y}_{t}$ represents the output value of the convolution at time step $t$, $M$ denotes the size of the convolutional kernel, $y_{t-m}$ represents the value of the interaction sequence $y$ at time step $t-m$, and $s_{m}$ is the weight of the convolutional kernel $s$ at position $m$. Then, $\tilde{y}_{t}$ goes through three residual blocks for feature extraction, and the fused learning state feature $\hat{y}_{t}$, which incorporates the learner's multi-scale learning patterns, is obtained by merging the output features of each residual block.
}
% 在这里，$\hat{y}_{t}$代表卷积在时间步$t$处的输出值，$M$表示卷积核的大小，$y_{t-m}$表示交互序列$y$在时间步$t-m$的值，$s_{m}$则是卷积核$s$在位置$m$的权重。之后$\tilde{y}_{t}$依次经过三个残差块进行特征提取，并通过融合每个残差块的输出特征得到融合了学习者多尺度学习模式的学习状态特征$\hat{y}_{t}$。

Through the LSE module, the model is able to capture the learner's learning patterns over multiple time scales during the answering process. The LSE module does not aggregate useful information through the similarity between exercises, but obtains a cross-knowledge concept learning pattern through the learner's recent performance. We believe that $\hat{y}_{t}$ can effectively describe the learner's learning state at time $t$.
% 通过LSE模块，模型能够捕获学习者在答题过程中多时间尺度下的学习模式。LSE模块不通过习题间的相似性聚合有效信息，而是通过学习者的近期表现获取一种跨知识概念的学习模式，我们认为$\hat{y}_{t}$能够有效地描述学习者在$t$时刻的学习状态。

}
\subsection{ Learning state enhanced knowledge state extraction module}
% 学习状态增强的知识状态提取
\par{
Through the LSE module, we obtained the learning state sequence of learners. However, the differences in the learner's states during the answering process is not involved in the process of extracting knowledge states. To take it into consideration, A learning state enhanced knowledge state extraction module was designed, as shown in \Cref{FIG:2}(c). 
%通过LSE模块，我们得到了学习者的学习状态序列。然而，学习者在回答过程中的状态差异不包括在提取知识状态的过程中。为此，我们设计了一个学习状态增强的知识状态提取模块，如图\Cref{FIG:2} (c)所示。

In the learner's exercise-answer interactions, the learning state may change due to various complex factors. However, not all past learning states are equally important for prediction. Due to the nature of the softmax function, even historical states with relatively small relevance to the current prediction may receive some attention, which could introduce additional noise in the process of extracting knowledge states. To address this issue, we adopted a sparse attention-weighted approach to incorporate consideration of the learning state into the process of extracting knowledge states.
% 在学习者的答题交互中，学习状态会因为各种复杂因素而发生变化。然而，并非所有过去的学习状态对预测都有同样的重要性。由于softmax函数的特性，即便是与当前预测关联性较小的历史状态也可能获得一定的关注度，这可能会在提取知识状态的过程中引入额外的噪声。为了解决这个问题，我们采用了稀疏注意力加权的方法，将对学习状态的考量融入到知识状态的提取过程中。

}
\par{
To exclude historically irrelevant moments,The $k$-means algorithm is used to group the historical learning state sequences, with different groups of historical moments being treated as irrelevant and masked. To obtain stably updated clustering clusters and ensure no leakage of future information, A fixed-size pool $X_{pool}$ with a size of $\mu$ is designed by us to store the learning states of the most recent $\mu$ learners (excluding learners from the current batch). Subsequently, by applying the $k$-means  algorithm to divide the learning states of all learners in the pool into $n$ groups, we obtain $n$ cluster centers for learning states $\left\{l_{1},\cdots,l_{i}, \cdots, l_{n}\right\}$, where $l_{n} \in \mathbb{R}^{1\times D}$ . Then, based on these $n$ cluster centers, the learning states of learners in the current batch are divided into clusters to obtain the corresponding cluster labels. The formula is as follows:
% 为了排除历史上不相关的时刻，采用k-means算法对历史学习状态序列进行分组，将不同组的历史时刻视为无关时刻并加以屏蔽。为了得到稳定更新的聚类簇，并确保不泄露未来信息，我们设计了一个固定大小为$\mu$的池$X_{pool}$，用于存储最近$\mu$位学习者的学习状态（不包含当前batch的学习者）。随后，通过k-means算法将池内所有学习者的学习状态划分成$n$组，得到$n$个学习状态簇中心$\left\{l_{1},\cdots,l_{i}, \cdots, l_{n}\right\}$，其中，$l_{n} \in \mathbb{R}^{1\times D}$ 。然后，根据这n个簇中心对当前batch的学习者学习状态进行簇的划分，从而获得相应的簇标签。其公式如下：
{\small
\begin{equation}
\label{deqn_ex14}
\operatorname{Label}\left(\hat{y}_{t}\right)=\arg \min _{i} \sqrt{\sum_{p=1}^{D}\left(\hat{y}_{t p}-l_{i p}\right)^{2}}
\end{equation}
}
\par{
In the above formula, $\hat{y}_{t}$ represents the learning state of the current learner at time step $t$. Using this formula, we can determine the cluster to which this learning state belongs, denoted as $\operatorname{Label}\left(\hat{y}_{t}\right)$.
% 在上述公式中，$\hat{y}_{t}$ 表示当前学习者在第 $t$ 个时刻的学习状态。通过该公式，我们可以确定该学习状态所属的簇，即 $\operatorname{Label}\left(\hat{y}_{t}\right)$。
}
}
\par{
Next, we will use the learning state information of the current batch of learners to update the pool.
% 接着，我们将使用当前批次学习者的学习状态信息来更新池：
{\small
\begin{equation}
X_{\text {pool }}=\operatorname{deque}\left(\left[X_{\text {pool }}, \hat{y}\right], \text { maxlen }=\mu\right)
\label{Eq:12}
\end{equation}
}
where $deque$ is a double-ended queue, and $\mu$ is the fixed size of the pool.The most recent $\mu$ learner learning state sequences are always retained for stable updates to the clustering center, earlier information will be popped out of the pool.
% 其中$deque$是一个双端队列，$\mu$为池的固定大小。我们始终保留最新的$\mu$位学习者学习状态序列用于稳定的更新聚类中心，早期信息将会被弹出池。
}
\par{
Once the cluster to which each learner belongs is ascertained, we begin to identify crucial historical moments by calculating the similarity of the historical learning state sequence at each time step. 
% 得知每位学习者所属的聚类簇后，我们开始通过计算历史学习状态序列每个时刻的相似性寻找关键历史时刻。
{\small
\begin{equation}
\label{deqn_ex14}
\beta_{t}=\frac{\hat{y}_{t} \hat{y}_{history}^{T}}{\sqrt{D}} \quad \hat{y}_{history}=\left\{\hat{y}_{1}, \cdots, \hat{y}_{t}\right\}
\end{equation}
}
where $\beta_{t}$ represents the similarity distribution calculated based on the learning state $\hat{y}_{t}$ at time $t$.
% 在上述式子中，$\beta_{t}$ 表示根据$t$时刻的学习状态$\hat{y}_{t}$计算得出的相似度分布。
}
\par{
By masking the attention scores of different clusters, we achieve sparse attention learning of states. Using $Mask(\cdot)$ to represent a masking operation that selects features within the same group, we can obtain a sparse similarity score matrix for learning state interactions along the sequence.
% 通过将不同聚类簇的注意力分数mask掉，我们实现了学习状态的稀疏注意力。用$Mask(\dot{})$代表一种挑选出同组特征的掩码操作，通过这一操作，我们可以获得序列上学习状态之间的稀疏相似度分数矩阵。
{\small
\begin{equation}
\label{deqn_ex14}
Mask\left(\beta_{t, \tau}\right)=\left\{\begin{array}{cc}
\beta_{t, \tau} & \text { if Label }\left(\hat{y}_{t}\right)=\operatorname{Label}\left(\hat{y}_{\tau}\right) \\
-\infty  & \text { Otherwise }
\end{array}\right.
\end{equation}
}
where $\text { Label }\left(\hat{y}_{t}\right)=\operatorname{Label}\left(\hat{y}_{\tau}\right)$ indicates that $\hat{y}_{t}$ and $\hat{y}_{\tau}$ belong to the same cluster, and attention scores between learning states in different clusters will be masked. $\beta_{t, \tau}$ represents the similarity score of $\hat{y}_{\tau}$ with respect to $\hat{y}_{t}$.
% 上述式子中$\text { Label }\left(\hat{y}_{t}\right)=\operatorname{Label}\left(\hat{y}_{\tau}\right)$表示$\hat{y}_{t}$和$\hat{y}_{\tau}$属于同一个聚类簇内，不在同一个簇内的注意力分数将会被屏蔽。$\beta_{t, \tau}$表示$\hat{y}_{\tau}$对于$\hat{y}_{t}$的相似度分数,。
}
\par{
Next, we merge the sparse similarity score matrix of learning states with its corresponding exercise similarity score matrix to achieve enhanced knowledge state capture from learning states. To prevent future information leakage, interaction sequences beyond time step $t$ should not be included in the calculation and thus need to be masked. The formula is as follows:
% 然后我们将学习状态的稀疏相似度分数矩阵与其对应的习题序列计算出来的习题相似度分数矩阵进行相加融合，从而实现学习状态增强的知识状态捕获。为了防止泄露未来信息，时间步长$t$时刻之后的交互序列不应该参与计算，所以需要将其mask起来。公式如下：
{\small
\begin{equation}
\label{deqn_ex14}
\gamma_{t, \tau} \mkern-2mu = \mkern-2mu
\begin{cases}
\frac{e^{q_{t} k_{\tau}^{T} / \sqrt{D}}}{\sum\limits_{\tau=1}^{N} e^{q_{t} k_{\tau}^{T} / \sqrt{D}}}\mkern-2mu + \mkern-2mu \frac{e^{Mask\left(\beta_{t, \tau}\right)}}{\sum\limits_{\tau=1}^{N} e^{Mask\left(\beta_{t, \tau}\right)}} & \mkern-10mu\text {if } \tau \in(1, t) \\
\phantom{aaaaaaaaa}0 & \mkern-10mu\text {if } \tau \in(t\mkern-3mu+\mkern-3mu1, N)
\end{cases}
\end{equation}
}
\par{
Here, the exercise feature $x_{t}$ at time $t$ is utilized by us as the query item $q_{t}$, and the exercise feature $x_{\tau}$ at time $\tau$ is taken as the key item $k_{\tau}$. Through the dot product operation, we obtain the attention distribution of the exercise sequence. Next, this attention distribution is integrated with the attention distribution of the learning state sequence, thereby incorporating the learner's state changes during the learning process, resulting in a comprehensive attention score distribution $\gamma$. The focus of this approach lies in emphasizing historical moments with similar learning states, further refining the modeling of the answering process, and successfully introducing the diversity of learning state changes into the model. This will help improve the performance of the model.
% 在上述公式中，我们使用$t$时刻的习题特征$x_{t}$作为查询项（$q_{t}$），并将$\tau$时刻的习题特征$x_{\tau}$作为键值项（$k_{\tau }$），我们通过点积运算获得习题序列的注意力分布。接下来，我们将这个注意力分布与学习状态序列的注意力分布进行融合，由此把学习者在学习过程中状态的变化考虑进来，从而得到完整的注意力分数分布$\gamma$。这种方法的重点在于，我们通过强调具有相似学习状态的历史时刻，进一步细化了对答题过程的建模，并成功地在模型中引入了学习状态变化的差异性，这将有助于提高模型的性能。
}
}
\par{
Key moment information from the historical interaction sequence can be extracted to obtain the final knowledge state by utilizing the complete attention score matrix:
% 通过利用完整的注意力分数矩阵，可以从历史交互序列中提取关键时刻信息，以获取最终的知识状态。
{\small
\begin{align}
h_{t}=\sum_{\tau=1}^{N} \gamma_{t, \tau} v_{\tau}
\end{align}
}
where $v_{\tau } \in \mathbb{R}^{1\times D }$ represents the interaction feature $y_{\tau}$ of the learner at time $\tau$ as the value, while $h_{t} \in \mathbb{R}^{1\times D }$ represents the final acquired knowledge state, which includes historical priors of learning performance.
% 其中，$v_{\tau } \in \mathbb{R}^{1\times D }$表示将在$\tau$时刻的学习者的交互特征$y_{\tau}$作为value，而$h_{t} \in \mathbb{R}^{1\times D }$代表最终获取的知识状态，即包含了学习状态考量的历史先验表现。
}
\par{
Subsequently, we fuse the knowledge state with the learning state, with the formula as follows:
% 随后，我们将知识状态与学习状态进行融合，其公式如下：
{\small
\begin{align}
z_{t}=\left[h_{t} \| \hat{y}_{t}\right]W_{7}
\end{align}
}
where $z_{t } \in \mathbb{R}^{1\times D }$ represents the fusion of two states, and $W_{7} \in \mathbb{R}^{2D\times D}$ is a learnable parameter matrix. We will utilize $z_{t }$ for the final prediction of the KT task.
% 其中，$z_{t } \in \mathbb{R}^{1\times D }$代表两种状态的融合,$W_{6} \in \mathbb{R}^{2D\times D}$是一个可学习的参数矩阵。。我们将利用$ z_{t }$进行最终KT任务的预测。
}
\subsection{Prediction}
\par{
The final step involves predicting the learner's response to the next exercise at the subsequent time step. The module's inputs comprise the comprehensive state feature $z_{t}$ at the current time step and the embedding vector $x_{t+1}$ of the subsequent exercise.
% 最后是预测学习者对下一时刻题目的答题情况。模块的输入包括当前时刻的综合状态特征$z_{t}$与下一问题的嵌入向量$x_{t+1}$.
{\small
\begin{align}
\hat{r}_{t+1}=\sigma\left(\left[z_{t} \| x_{t+1}\right] W_{8}\right)
\end{align}
}
\par{Here, $\sigma$ is the Sigmoid function, and $W_{8} \in \mathbb{R}^{D\times D }$ is a learnable parameter matrix. These inputs are fused through a fully connected network, and ultimately, a sigmoid function is applied to generate the predicted probability $\hat{r}_{t+1}$ of the learner answering the current exercise correctly, where $\hat{r}_{t+1} \in \left[0,1\right]$.
% 这里，$\sigma$是Sigmoid函数，$W_{8} \in \mathbb{R}^{D\times D }$是一个可学习的参数矩阵。这些输入经过一个全连接网络融合，最后通过sigmoid函数生成学习者在当前问题上正确回答的预测概率$\hat{r}_{t+1} \in \left[0,1\right]$。
}
{\small
\begin{align}
Loss\mkern-2mu=\mkern-2mu-\mkern-8mu\sum_{t=1}\mkern-4mu\left(r_{t+1} \log \hat{r}_{t+1}\mkern-2mu+\mkern-3mu\left(\mkern-3mu1\mkern-2mu-\mkern-2mur_{t+1}\right) \log \left(\mkern-3mu1\mkern-2mu-\mkern-2mu\hat{r}_{t+1}\right)\right)
\end{align}
}
\par{
In the entire LSKT method, all learnable parameters are trained in an end-to-end manner by minimizing the binary cross-entropy loss of all learner responses.
% 在整个$LSKT$方法中，所有可学习的参数都以端到端方式通过最小化所有学习者响应的二元交叉熵损失进行训练。
}
}
\section{EXPERIMENTAL RESULTS}
\par{
In this section, a detailed presentation of the experimental results of LSKT on four real-world educational datasets is provided, accompanied by extensive discussions and analysis for the evaluation of the model performance. We address the following research questions:
% 在本节中，详细介绍了在四个真实世界教育数据集上的LSKT实验结果，并进行了广泛的讨论和分析，以评估模型性能。我们回答了以下研究问题：
}
\begin{itemize}
\item{\textbf{RQ1.} How does LSKT perform compared to baseline methods in predicting learners' future performance?} \vspace{-\topsep}
\item{\textbf{RQ2.} What is the role of the key model components of LSKT in the entire method framework?} \vspace{-\topsep}
\item{\textbf{RQ3.} How does the learning state promote more fine-grained knowledge state extraction?} \vspace{-\topsep}
\item{\textbf{RQ4.} What impact does the feature modeling of three granularities based on IRT have on model embedding?} \vspace{-\topsep}
\end{itemize}
% •RQ1。与基线方法相比，LSKT在预测学习者未来表现方面表现如何?
% •RQ2。LSKT的关键模型组件在整个方法框架中的作用是什么?
% •RQ3。学习状态如何促进更细粒度的知识状态提取?
% •RQ4。基于IRT的三种粒度的特征建模对模型嵌入的影响是什么?

\setcounter{footnote}{0} % 强制设置脚注的起始计数值为0。
\subsection{Datasets}
\par{
    To evaluate the performance of our LSKT model, we conducted experiments on four real-world publicly available datasets, including ASSIST09\footnote{\href{https://sites.google.com/site/assistmentsdata/home/2009-2010-assistment-data}{ https://sites.google.com/site/assistmentsdata/home/2009-2010-assistment-data}}, ASSIST12\footnote{\href{https://sites.google.com/site/assistmentsdata/datasets/2012-13-school-data-with-affect}{https://sites.google.com/site/assistmentsdata/datasets/2012-13-school-data-with-affect}}, ASSISTChall\footnote{\href{https://sites.google.com/view/assistmentsdatamining/}{https://sites.google.com/view/assistmentsdatamining/}}, and algebra05\footnote{\href{https://pslcdatashop.web.cmu.edu/KDDCup/}{https://pslcdatashop.web.cmu.edu/KDDCup/}}. The ASSIST09 dataset~\citep{[18]}  was created and collected by the online tutoring system ASSISTment in 2004. Another dataset from the same platform is ASSIST12, which collected data from 2012 to 2013 and has been regarded as one of the benchmark datasets for KT research over the past decade. The ASSISTChall~\citep{[19]}  dataset was released in a data mining competition in 2017, containing longer sequences of learner interactions and allowing multiple attempts for a single problem. The algebra0~\citep{[20]}  dataset was released in the KDDcup 2010 Educational Data Mining Challenge, comprising learner responses to algebra problems from 2005 to 2006.
    % 为了评估我们的LSKT模型的性能，我们在四个真实世界的公开数据集上进行了实验，包括ASSIST09、ASSIST12、ASSISTChall和algebra05。ASSIST09[18]数据集是由在线辅导系统ASSISTment在2004年创建并收集的。同样来自该平台的另一个数据集是ASSIST12，它收集了2012年至2013年的数据，这个数据集在过去十年中一直被视为KT研究的基准数据集之一。ASSISTChall[19]数据集是在2017年的数据挖掘竞赛中发布的，它包含了学习者更长的学习序列，并允许对一个问题进行多次尝试。algebra05[20]数据集是在KDDcup 2010教育数据挖掘挑战赛上发布的，它收录了2005年至2006年学习者对代数问题的回答。
    
    For fairness, these datasets were preprocessed based on prior research~\citep{[14]}, with data modeling issues being addressed and duplicate records being removed. As the ASSIST09 dataset lacks timestamp information, we followed previous studies~\citep{[38],[39]} and sorted learners' answering sequences based on order ID. The detailed statistical results of the processed datasets are presented in \Cref{Table:2}.
% 我们根据之前的研究[14]对这些数据集进行了预处理，修复了数据建模问题，并删除了重复记录。由于ASSIST09数据集缺少时间戳信息，我们参考了先前的研究[38][39]，根据order id对学习者的答题序列进行了排序。经过处理后的数据集的详细统计结果如表2所示。
}

\begin{table}[]
\centering
\caption{Dataset statistics. }
\label{Table:2}
\footnotesize
\setlength\tabcolsep{5pt}
\begin{tabular}{@{}lllll@{}}
\toprule
Statistics         & ASSIST09 & ASSIST12  & ASSISTChall & algebra05 \\ \midrule
Interactions       & 346,860  & 2,711,813 & 942,816     & 809,694   \\
Learners           & 4,217    & 29,018    & 1709        & 574       \\
Exercises          & 26,688   & 53,091    & 3,162       & 1,084     \\
Concepts & 123      & 265       & 102         & 138       \\
Avg.Length         & 82.25    & 93.45     & 873.79      & 1,410.62  \\ \bottomrule
\end{tabular}
\end{table}
\subsection{Baseline methods}
\par{
To evaluate the effectiveness of LSKT, this section will select several classic or state-of-the-art methods as baseline methods, such as DKT~\citep{[11]}, DKVMN~\citep{[12]}, SAKT~\citep{[13]}, AKT~\citep{[2]}, DTransformer~\citep{[14]}, and FKT~\citep{[37]}.
% 为了评估LSKT的有效性，本节将选择介绍几个最经典或最先进的方法作为基线方法，如DKT~\citep{[11]}、DKVMN~\citep{[12]}、SAKT~\citep{[13]}、AKT~\citep{[2]}、DTransformer~\citep{[14]}和 FKT~\citep{[37]}。
}

\par{
Among them, DKT was a milestone method that introduced recursive neural networks into the field of knowledge tracing, and it outperformed traditional knowledge tracing models. DKVMN further advanced the development of KT by storing knowledge state in a dynamic key-value memory network and updating it at different interaction moments.CKT attempted for the first time to use Convolutional Neural Networks (CNN) to perform knowledge tracing prediction tasks, simulating personalized learning rates for each learner through convolutional operations. SAKT applied transformer encoders to KT and proposed a self-attention mechanism-based method, treating the target problem as a query and the history exercise-answer pairs as keys and values. AKT continued this approach by encoding knowledge state and exercises using self-attention neural networks, and then utilizing a knowledge retriever to retrieve future knowledge states. DTransformer was built upon AKT and designed a knowledge-level diagnostic extractor that could explicitly diagnose learners' proficiency levels. Contrastive learning was introduced during training to maintain the stability of knowledge state. FKT divided the KT prediction task into three steps: obtaining historical knowledge state, inferring future latent features, and predicting future performance. It developed an encoder-decoder-predictor framework to further improve the accuracy of knowledge tracing.
% 其中，DKT是首个将递归神经网络引入知识追踪领域的里程碑式方法，其表现优于传统的知识追踪模型。DKVMN进一步推动了KT领域的发展，其通过动态键值记忆网络存储知识状态，并在不同的交互时刻更新知识状态。CKT首次尝试利用卷积神经网络（CNN）完成知识追踪预测任务，通过卷积操作来模拟每个学习者的个性化学习率。SAKT将transformer编码器应用于KT，并提出了一种基于自关注机制的方法，它将目标问题视为查询，将历史问题-答案对视为键和值。AKT继续发展这一思路，利用自关注神经网络对知识状态和问题进行编码，然后利用知识检索器进行检索，以获得未来的知识状态。Dtransformer在AKT的基础上，设计了一种知识级诊断提取器，能够明确地诊断出学习者的知识熟练程度，并在训练中引入对比学习，保持了知识状态的稳定性。FKT将KT预测任务分为三个步骤:获取历史知识状态、推断未来潜在特征和预测未来表现,并开发了一个编码器-解码器-预测器框架，进一步提升了知识追踪的精度。
}

\subsection{Implementation details}
\par{
In our experiment, following the data preprocessing method from previous work ~\citep{[14]}, we split the learning records of 80$\%$ of the learners as the training set and 20$\%$ as the test set. For all data, the learners' learning records were first sorted according to the timestamp, then, as per work~\citep{[2],[14]}, learner response sequences with a length exceeding 200 were truncated. For shorter sequences, zero padding was employed to pad them to a fixed length of 200, a process that aids in improving computational efficiency.
% 在我们的实验中，根据之前工作[14]的数据集处理方法，我们将80%的学习者的学习记录作为训练集，20%的学习者的学习记录作为测试集。对于所有数据，先根据时间戳对学习者的学习记录进行排序，然后根据工作[2,14]，截断长度超过200的学习者响应序列。对于较短的序列，使用零填充将他们填充到固定长度200，这样处理有助于提高计算效率。
}
\par{
The AdamW optimizer with a learning rate of 0.001 and a batch size of 16 is leveraged to train the model, where the dimensions of the exercise embeddings, interaction embeddings, and the answer embeddings are all set to 128. For the hyperparameters, we set the parameter $M$ in \Cref{Eq:10} to 3 and the parameter $\mu$ in \Cref{Eq:12} to the batch size 16. After adjusting the number of clustering clusters $n$ from 1 to 10, we set the value of n to 4. For fair comparison, all models have been fine-tuned to achieve the best performance. We set the threshold for predicting the response answer to 0.5~\citep{[22],[23],[24]}, where responses with a probability greater than 0.5 are considered correct answers; otherwise, they are considered incorrect answers. Then, the accuracy of predicting response answers was evaluated using area under the curve~(AUC), accuracy~(ACC), root mean square error~(RMSE), and mean absolute error~(MAE).
% 我们使用学习率为0.001的AdamW优化器，并使用批量大小为16来训练模型，其中问题嵌入、交互嵌入和答案嵌入的维度均设置为128。对于超参数，我们将\Cref{Eq:10}中的参数M设置为3，将\Cref{Eq:12}中的参数$\mu$设置为批量大小16。在将聚类簇的数量n从1调整到10后，我们将n的值设置为4。为了公平比较，所有模型都经过了调优以达到最佳性能。我们将预测响应答案的阈值设置为0.5~\citep{[22],[23],[24]}，其中概率大于0.5的响应被视为正确答案；否则，它们被视为不正确答案。然后，我们使用曲线下面积（AUC）、准确率（ACC）、均方根误差（RMSE）和平均绝对误差（MAE）来评估预测响应答案的准确性。
}

\begin{table*}[width=2\linewidth]
\caption{Comparison results with baseline methods on the AUC and ACC metrics.The best and the second-best results are marked in boldface and underlined, respectively.}
\label{Table:3}
\renewcommand{\arraystretch}{1.14}
\begin{tabular}{llccccccccc}
\hline
\multirow{2}{*}{Dataset}     & \multirow{2}{*}{Metric} & \multicolumn{7}{c}{Baseline}                                             & \multirow{2}{*}{LSKT-NI} & \multirow{2}{*}{LSKT} \\ \cline{3-9}
                             &                         & DKT    & DKVMN  & CKT    & SAKT   & AKT    & DTransformer & FKT          &                          &                       \\ \hline
\multirow{4}{*}{ASSIST09}    & AUC                     & 0.7908 & 0.7891 & 0.8072 & 0.7783 & 0.8162 & 0.8171       & —            & {\ul 0.8206}             & \textbf{0.8369}       \\
                             & ACC                     & 0.6482 & 0.7458 & 0.7622 & 0.7418 & 0.7639 & {\ul 0.7655} & -            & 0.7650                   & \textbf{0.7785}       \\
                             & RMSE                    & 0.4387 & 0.4161 & 0.4089 & 0.4194 & 0.4065 & 0.4005       & -            & {\ul 0.3966}             & \textbf{0.3886}       \\
                             & MAE                     & 0.4105 & 0.3132 & 0.3121 & 0.3342 & 0.3109 & 0.3069       & -            & {\ul 0.3043}             & \textbf{0.2896}       \\ \hline
\multirow{4}{*}{ASSIST12}    & AUC                     & 0.7060 & 0.7137 & 0.7310 & 0.7021 & 0.7632 & 0.7598       & {\ul 0.7692} & 0.7348                   & \textbf{0.7781}       \\
                             & ACC                     & 0.6998 & 0.7295 & 0.7365 & 0.7232 & 0.7415 & 0.7392       & {\ul 0.7485} & 0.7339                   & \textbf{0.7559}       \\
                             & RMSE                    & 0.4456 & 0.4422 & 0.4234 & 0.4420 & 0.4199 & 0.4223       & {\ul 0.4144} & 0.4243                   & \textbf{0.4098}       \\
                             & MAE                     & 0.3655 & 0.3634 & 0.3612 & 0.3663 & 0.3476 & 0.3565       & {\ul 0.3321} & 0.3585                   & \textbf{0.3139}       \\ \hline
\multirow{4}{*}{ASSISTChall} & AUC                     & 0.6809 & 0.7024 & 0.7262 & 0.6726 & 0.7556 & 0.7512       & {\ul 0.7584} & 0.7244                   & \textbf{0.7917}       \\
                             & ACC                     & 0.6971 & 0.6775 & 0.6924 & 0.6735 & 0.7112 & 0.7052       & {\ul 0.7133} & 0.6896                   & \textbf{0.7341}       \\
                             & RMSE                    & 0.4543 & 0.4543 & 0.4455 & 0.4609 & 0.4355 & 0.4389       & {\ul 0.4311} & 0.4474                   & \textbf{0.4197}       \\
                             & MAE                     & 0.4143 & 0.4064 & 0.3864 & 0.4207 & 0.3733 & 0.3767       & {\ul 0.3695} & 0.3967                   & \textbf{0.3451}       \\ \hline
\multirow{4}{*}{algebra05}   & AUC                     & 0.7558 & 0.7857 & 0.7899 & 0.7819 & 0.7966 & 0.7932       & {\ul 0.7981} & 0.7969                   & \textbf{0.8063}       \\
                             & ACC                     & 0.8263 & 0.8315 & 0.8384 & 0.8337 & 0.8414 & 0.8406       & {\ul 0.8421} & 0.8412                   & \textbf{0.8448}       \\
                             & RMSE                    & 0.3798 & 0.3424 & 0.3417 & 0.3465 & 0.3406 & 0.3415       & {\ul 0.3394} & 0.3398                   & \textbf{0.3360}       \\
                             & MAE                     & 0.3498 & 0.2315 & 0.2281 & 0.2304 & 0.2252 & {\ul 0.2221} & 0.2266       & 0.2420                   & \textbf{0.2211}       \\ \hline
\end{tabular}
\end{table*}

\subsection{Performance prediction (RQ1)}
\par{
As shown in \Cref{Table:3}, to evaluate the predictive performance of the LSKT model, we compared it with seven methods in the KT domain. Among them, LSKT-NI indicates not using three embedding methods based on the IRT model, but directly utilizing knowledge concept embedding features to train the model. Noeworthily, due to the the training of the FKT model requires the use of timestamp labels to obtain response times, and the ASSIST09 dataset lacks such timestamp label information, we only conducted experiments on the FKT model on the other three datasets excluding ASSIST09.
% 如表3所示，我们为了评估LSKT模型的预测性能，将其与KT领域中的七种方法进行了对比。其中LSKT-NI表示不使用基于IRT模型的三种嵌入方法，而是直接利用知识概念嵌入来训练模型。需要指出的是，由于FKT模型的训练需要利用时间戳标签来获取响应时间，而ASSIST09数据集中缺乏这样的时间戳标签信息，因此我们仅在除了ASSIST09以外的其他三个数据集上对FKT模型进行了实验。
}
\par{
From~\Cref{Table:3}, we can see that the proposed model achieves superior performance across four different datasets. Specifically, compared with the current state-of-the-art methods, our model shows an improvement of 1.98$\%$ in AUC on the ASSIST09 dataset, 0.89$\%$ on the ASSIST12 dataset, 3.33$\%$ on the ASSISTChall dataset, and 0.82$\%$ on the algebra05 dataset. Besides excelling in the AUC metric, our model also exhibits advantages in other performance evaluation metrics, further proving its robust capability in performing knowledge tracing tasks. However, if we do not adopt a strategy to model the differences between interactions but directly use knowledge concept embeddings, the performance of the model on these four datasets will significantly decline. It is noteworthy that baseline models based on Transformers, such as AKT, DTransformer, and FKT, all to model the differences between exercises to some degree, which is also one of the reasons for their performance superiority over SAKT. All of these highlight the importance of modeling exercise differences for knowledge tracing tasks, especially for the ATT-DLKT model. In RQ2, more detailed ablation experiment results will be further discussed.
% 从表格中我们可以看出，我们提出的模型在四个不同的数据集上都表现出了最优秀的性能。具体说来，与当前最先进的方法相比，我们的模型在ASSIST09数据集上的AUC提升了1.98%，在ASSIST12数据集上提升了0.89%，在ASSISTChall数据集上提升了3.33%，在algebra05数据集上则提升了0.82%。除了在AUC这个指标上的优异表现外，我们的模型在其他性能评价指标上也展现出了优势，这进一步证明了我们的模型在执行知识追踪任务时的强大能力。然而，如果我们不采取对题目间差异性进行建模的策略，而是直接使用知识概念嵌入，模型在这四个数据集上的性能会有明显的下滑。值得注意的是，基于Transformer的Baseline模型，如AKT，DTransformer和FKT，都在不同程度上对题目差异性进行了建模，这也是它们性能优于SAKT的一个原因。所有这些都突显了习题差异性建模对知识追踪任务，尤其是对于ATT-DLKT模型的重要性。在RQ2中，我们将进一步讨论并分析更详细的消融实验结果。
}
\begin{table}[width=1\linewidth]
{\footnotesize
\centering
\begin{threeparttable}
\caption{ Comparison of AUC and ACC among three embedding modeling methods.The best and the second-best results are marked in boldface and underlined, respectively.}
\label{Table:4}
\renewcommand{\arraystretch}{1.3}
\setlength{\tabcolsep}{5pt}
\begin{tabular}{llccc}
\hline
Dataset                      & Metric & LSKT-1PL     & LSKT-2PL        & LSKT-3PL        \\ \hline
\multirow{2}{*}{ASSIST09}    & AUC    & 0.8217       & {\ul 0.8289}    & \textbf{0.8369} \\
                             & ACC    & 0.7669       & {\ul 0.7740}    & \textbf{0.7785} \\ \hline
\multirow{2}{*}{ASSIST12}    & AUC    & 0.7754    & {\ul 0.7769}          & \textbf{0.7781} \\
                             & ACC    & 0.7530     & {\ul 0.7532}     & \textbf{0.7559} \\ \hline
\multirow{2}{*}{ASSISTChall} & AUC    & 0.7776       & {\ul 0.7904}    & \textbf{0.7917} \\
                             & ACC    & 0.7220       & {\ul 0.7336}    & \textbf{0.7341} \\ \hline
\multirow{2}{*}{algebra05}   & AUC    & 0.7982       & \textbf{0.8096} & {\ul 0.8063}    \\
                             & ACC    & 0.8455       & \textbf{0.8471} & {\ul 0.8448}    \\ \hline
\end{tabular} 
\end{threeparttable}
}
\vspace{-0.5em}
\end{table}
\subsection{Ablation experiments (RQ2)}
\par{
In order to investigate the impact of each key component in LSKT on the model's predictive results, we further conducted ablation experiments.
% 为了讨论LSKT中每个关键组件对于模型预测结果的影响，我们进一步进行了消融实验。
}
\par{
We first compared the effects of three different granularity levels of differential embedding modeling on the performance of the knowledge tracing model. As shown in \Cref{Table:4}, we employed three embedding modeling methods: LSKT-1PL, LSKT-2PL, and LSKT-3PL. The LSKT-1PL model simulates the differences between exercises by introducing difficulty distinctions among exercises into the modeling of exercise features and interaction features, thus modeling the differences between exercises at a coarse granularity. LSKT-2PL, building on the introduction of difficulty distinctions, further models the differentiation between exercises on the same knowledge concept by utilizing a higher-dimensional exercise mapping feature, achieving a sub-fine-grained modeling of the differences between exercises. The LSKT-3PL model introduces a random guessing factor based on LSKT-2PL, enabling a finer-grained simulation of the complex learning behaviors of learners during the answering process. Experimental results show that the LSKT-3PL model achieves the best performance on the ASSIST09, ASSIST12, and ASSISTChall datasets, while its performance on the algebra05 dataset is suboptimal. This suggests that for datasets with fewer exercise types such as algebra05, modeling the differences between exercises at a finer granularity does not necessarily lead to better predictive performance. Perhaps the sub-fine-grained modeling of the LSKT-2PL embedding modeling is sufficient to simulate the differences between exercises on the same concept. Therefore, we compared these three different granularity levels of differential modeling to find a balance between model performance and granularity of modeling. In RQ4, the differences between these three modeling approaches and their impact on embedding features will be further analyzed.
% 我们首先对比了三种不同粒度的练习差异度嵌入建模对知识追踪模型效果的影响。如表4所示，我们使用了LSKT-1PL、LSKT-2PL和LSKT-3PL三种嵌入建模方法。LSKT-1PL模型通过将习题间的难度区分引入习题特征与交互特征的建模，粗粒度的模拟了习题之间的差异。LSKT-2PL在引入困难度区分的基础上通过利用一个更高维的习题映射特征进一步模拟同知识概念习题间的区分度差异，从而实现次细粒度的习题间差异建模。LSKT-3PL模型在LSKT-2PL的基础上引入了随机猜测因素，从而能够更细粒度地模拟学习者答题过程中的复杂学习行为。实验结果显示，LSKT-3PL模型在ASSIST09，ASSIST12和ASSISTChall数据集上均取得了最佳效果，而在algebra05数据集上的效果则为次优。这表明对于algebra05等习题种类较少的数据集，并不是习题间的差异性建模越细粒度模型预测效果就越好，也许次细粒度的LSKT-2PL嵌入建模可能就已经足够模拟同概念习题之间的差异。为此，我们对比了这三种不同粒度的差异性建模，目的是为了寻找模型效果与建模粒度之间的平衡。在RQ4中将对这三种建模之间的差异，以及它们对嵌入特征的影响将得到进一步的分析。
}

\begin{figure*}
\begin{minipage}[b]{0.23\textwidth}
    \hspace{1.5em}
    \includegraphics[height=0.1\linewidth]{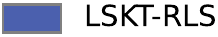}
  \end{minipage}
  \begin{minipage}[b]{0.23\textwidth}
  \centering
    % \hspace{-2.9em}
    \includegraphics[height=0.1\linewidth]{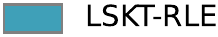}
  \end{minipage}
  \begin{minipage}[b]{0.23\textwidth}
  \centering
    % \hspace{-2.9em}
    \includegraphics[height=0.1\linewidth]{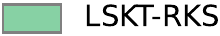}
  \end{minipage}
  \begin{minipage}[b]{0.23\textwidth}
  \centering
    % \hspace{-2.9em}
    \includegraphics[height=0.1\linewidth]{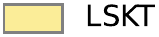}
  \end{minipage}
% \begin{minipage}[b]{1\textwidth}
%     \hspace{3em}
%     \includegraphics[height=0.04\linewidth,width = 0.92\linewidth]{coler_lable.pdf}
%     % \vspace{1em}
%   \end{minipage}
  
  \vspace{0.4em}
\begin{minipage}[b]{0.2529\textwidth}
    \hspace{-5.5em}
    \includegraphics[height=1\linewidth]{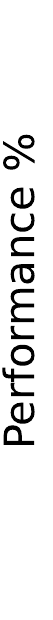}
    \vspace{2em}
  \end{minipage}\hspace{-15.5em}
  \begin{minipage}[b]{0.3\textwidth}
  \hspace{-4em}
    \centering
    \includegraphics[height=0.9\linewidth,width=0.98\linewidth]{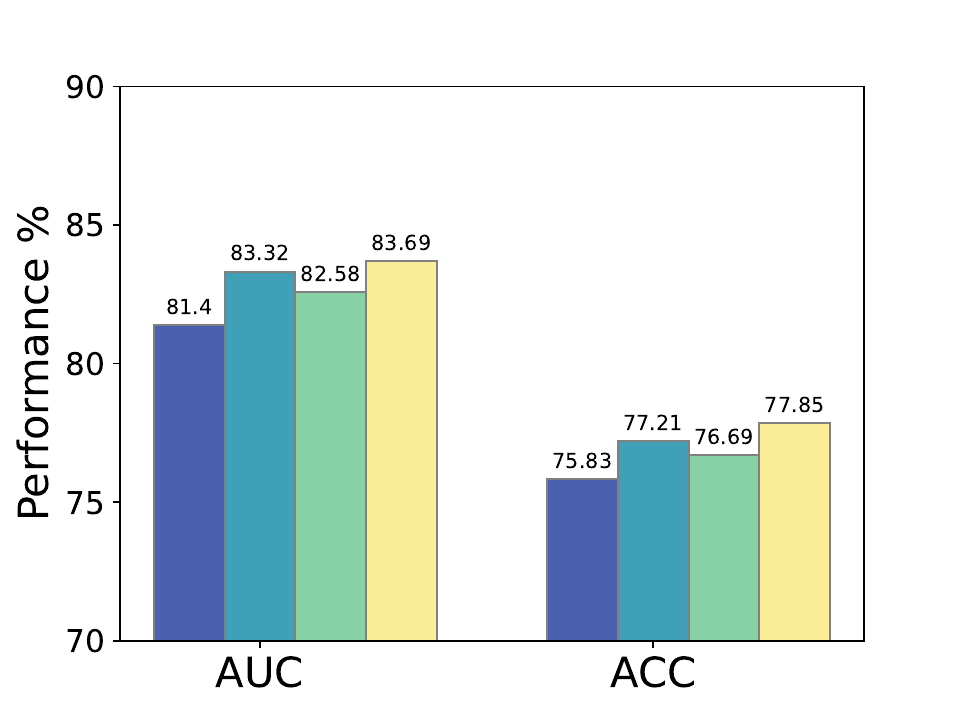}
    ASSIST09~~~~~~~~
  \end{minipage}\hspace{-0.2em}
  \begin{minipage}[b]{0.3\textwidth}
  \hspace{-3.2em}
    \centering
    \includegraphics[height=0.9\linewidth,width=0.98\linewidth]{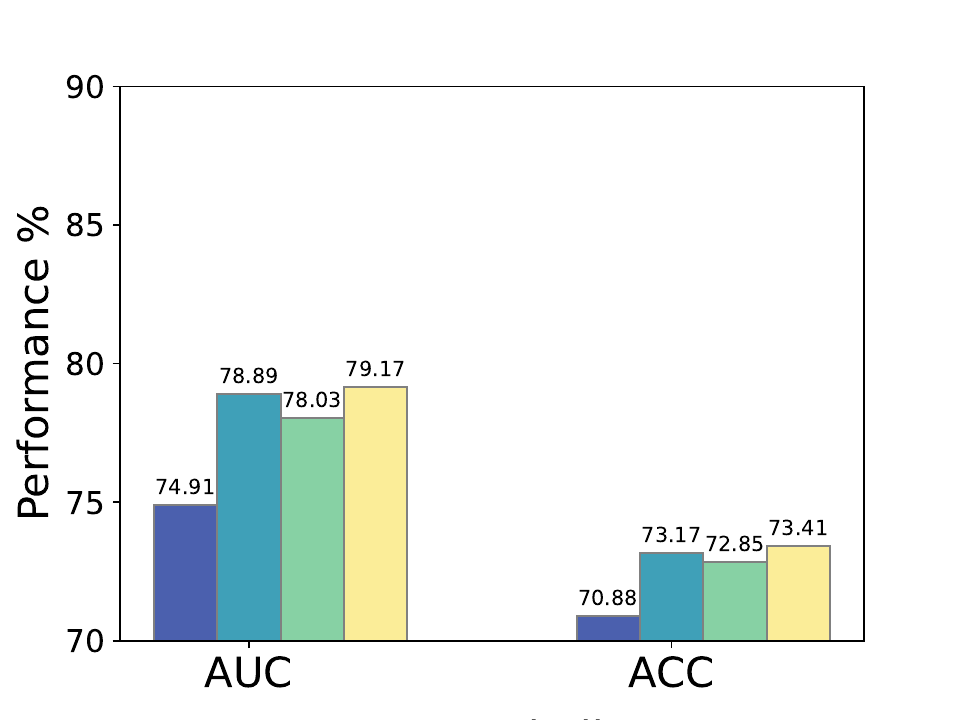}
    ASSISTChall ~~~~
  \end{minipage}\hspace{-0.2em}
  \begin{minipage}[b]{0.3\textwidth}
    \hspace{-2.8em}
    \centering
    \includegraphics[height=0.9\linewidth,width=0.98\linewidth]{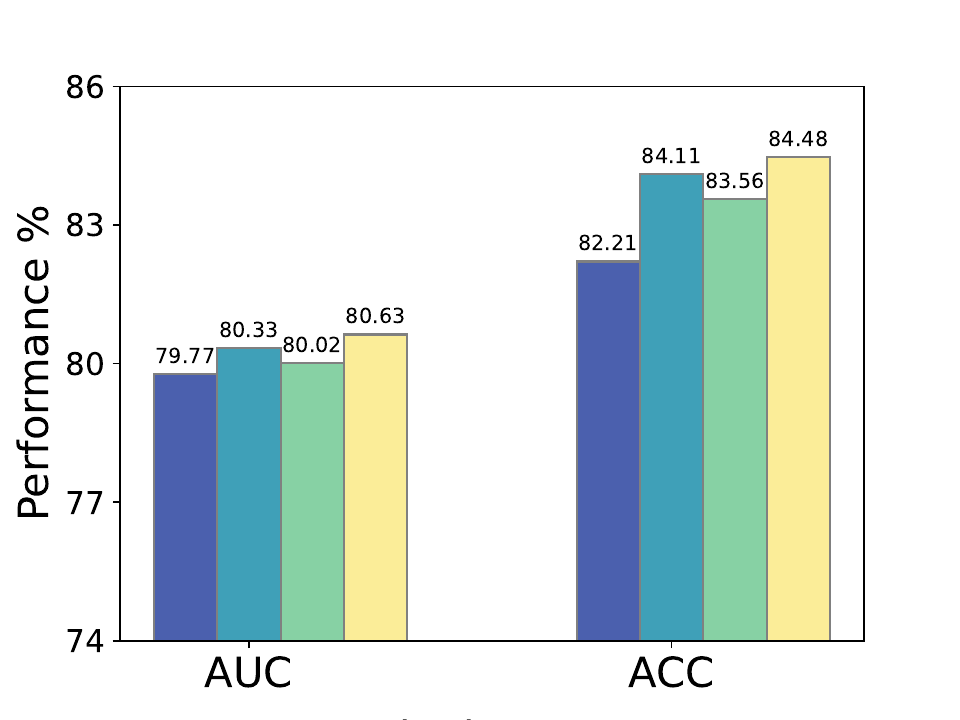}
    algebra05~~~~
  \end{minipage}
\caption{Comparison of ablation experiment results on three datasets. Different colors are used to distinguish between different ablation models, and the specific experimental results are labeled at the top of each bar chart.}
% 在三个代表数据集上的消融实验对比结果。使用不同的颜色标记不同的消融模型，具体实验结果被标记在对应柱形的顶部。
  \label{FIG:3}
\vspace{-1em}
\end{figure*}

\par{
Next, we will analyze the effectiveness of various key components of the LSKT model. We compared the differences in ablation study in \Cref{Table:5} and the ablation results on the ASSIST09, ASSISTChall, and algebra05 datasets are shown in \Cref{FIG:3}.
% 接下来，我们将分析LSKT模型各个关键组成模块的有效性。我们在表5中比较了消融设置的差异，并在图3中展示了在ASSIST09，ASSISTChall和algebra05三个数据集上的消融结果。
}
\begin{itemize}
\item{RLS (i.e., Removal of Learning State Extraction Module) disregards the impact of changes in learning state on the model.} \vspace{-\topsep}
\item{RLE (i.e., Removal of Learning State Enhancement) retains both learning state and knowledge state for model prediction but does not consider the influence of learning state on acquiring knowledge state.} \vspace{-\topsep}
\item{RKS (i.e., Removal of Knowledge State Extraction Module) disregards capturing knowledge state from the history of responses and only utilizes learning state for prediction tasks.} \vspace{-\topsep}
\end{itemize}
% -RLS（即删除学习状态提取模块）不考虑学习状态的变化对模型产生的影响。
% -RLE (即不使用学习状态增强知识状态的提取)保留学习状态与知识状态共同参与模型预测，但是不考虑学习状态对获取知识状态的影响。
% -RKS （即删除知识状态提取模块）不考虑从答题历史序列中捕知识状态，仅使用学习状态进行预测任务。

\begin{table}[]
{\footnotesize
\centering
\renewcommand{\arraystretch}{0.65} 
\caption{Different variants of comparative settings.}
\label{Table:5}
\begin{tabular}{lccc}
\hline
Methods                   & \begin{tabular}[c]{@{}c@{}}Learning state \\ extraction\end{tabular} & \begin{tabular}[c]{@{}c@{}}Learning state\\ enhancement\end{tabular} & \begin{tabular}[c]{@{}c@{}} knowledge
state \\ extraction\end{tabular} \\ \hline
\multirow{3}{*}{LSKT-RLS}   & \multirow{3}{*}{\usym{2715}}                                                          & \multirow{3}{*}{\usym{2715}}                                                    & \multirow{3}{*}{\usym{2713}} \\
                            &                                                                             &                                                                       &                    \\
                            &                                                                             &                                                                       &                    \\
\multirow{3}{*}{LSKT-RLE}   & \multirow{3}{*}{\usym{2713}}                                                          & \multirow{3}{*}{\usym{2715}}                                                    & \multirow{3}{*}{\usym{2713}} \\
                            &                                                                             &                                                                       &                    \\
                            &                                                                             &                                                                       &                    \\
\multirow{3}{*}{LSKT-RKS}   & \multirow{3}{*}{\usym{2713}}                                                          & \multirow{3}{*}{\usym{2715}}                                                    & \multirow{3}{*}{\usym{2715}} \\
                            &                                                                             &                                                                       &                    \\
                            &                                                                             &                                                                       &                    \\
\multirow{3}{*}{LSKT}       & \multirow{3}{*}{\usym{2713}}                                                          & \multirow{3}{*}{\usym{2713}}                                                    & \multirow{3}{*}{\usym{2713}} \\
                            &                                                                             &                                                                       &                    \\
                            &                                                                             &                                                                       &                    \\ \hline
\end{tabular}
}
\vspace{-1em}
\end{table}

\par{
 The experimental results are displayed in \Cref{FIG:3}. It can be observed that, firstly, the absence of either the knowledge state or the learning state leads to a decrease in model performance. Only by comprehensively considering both factors can the model's performance reach its optimum. This indicates that the knowledge state and the learning state can complement each other well. Secondly, introducing the distinction of learner state during the extraction of knowledge state can also enhance the model's performance. It is noteworthy that the features of the learning state have a more significant impact on the model's predictive effect. This phenomenon may be due to the influence of forgetting factors in human's actual learning process, which makes the learner's performance at any given moment more affected by recent learning interactions rather than the cumulative interactions of a long history. This explains why the AUC of the LSKT-RKS model is generally superior to that of the LSKT-RLS model on all datasets, a finding that meets our perception.
% 实验结果展示在图3中。可以观察到，首先，知识状态的的缺失或者是学习状态的缺失都会导致模型效果的下降，只有综合考虑两者，模型的效果才会达到最优。这说明知识状态和学习状态这两个因素能够很好的互补。其次，在知识状态的提取过程中引入学习者状态的区分也可以提升模型的性能。值得关注的是，学习状态的特征对模型预测效果的影响更为显著。存在这一现象可是因为人类的真实学习过程中存在遗忘因素的影响，这使得学习者在任何给定时刻的表现更多的受到近期学习交互，而不是长期历史的累积交互的影响。这解释了为什么在所有数据集上，LSKT-RKS模型的AUC要优于LSKT-RLS模型，这一发现符合我们的直觉。
}

\begin{figure*}
  \begin{minipage}[b]{0.47\textwidth}
    \centering
    \hspace{-2em}
    \includegraphics[width=0.83\linewidth]{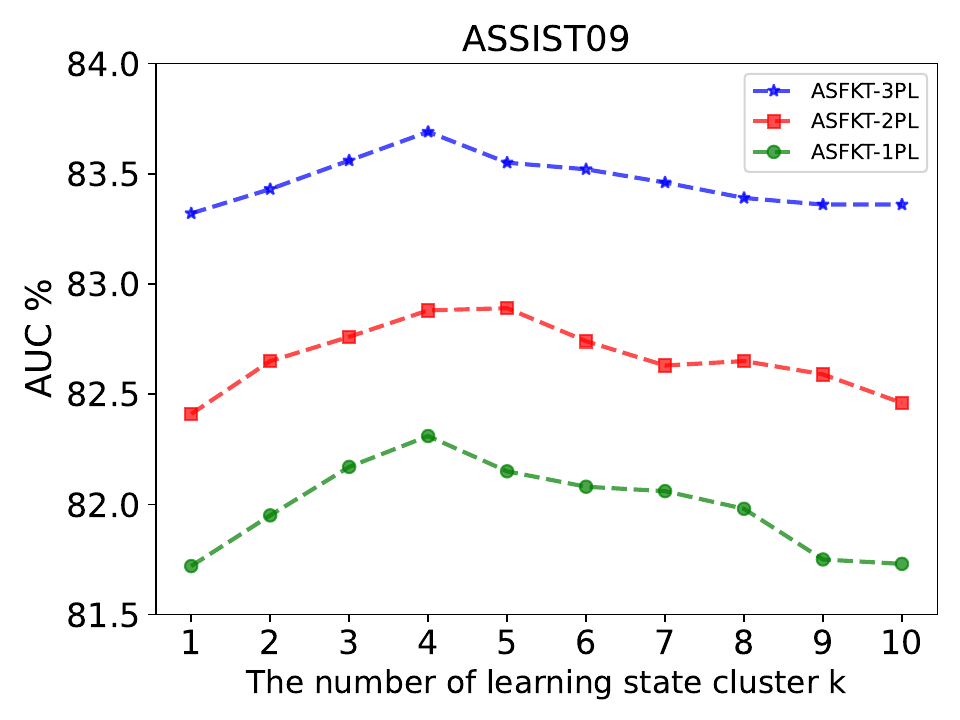}
  \end{minipage}
  \begin{minipage}[b]{0.47\textwidth}
  \hspace{-3.5em}
    \includegraphics[width=0.83\linewidth]{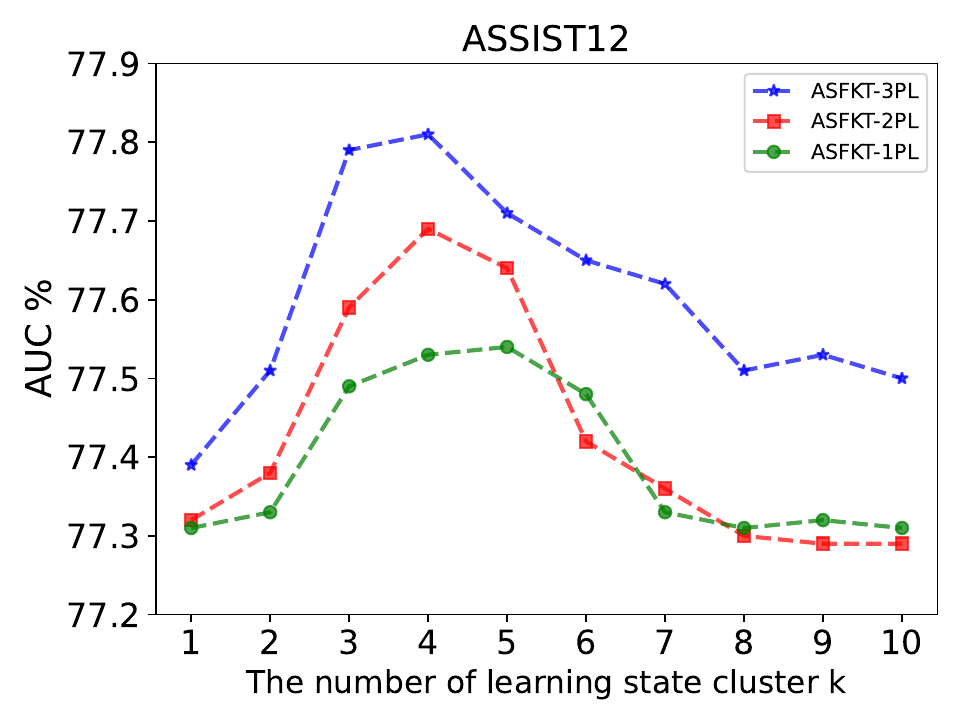}
  \end{minipage}

  \begin{minipage}[b]{0.47\textwidth}
    \centering
    \hspace{-5.8em}
    \includegraphics[width=0.83\linewidth]{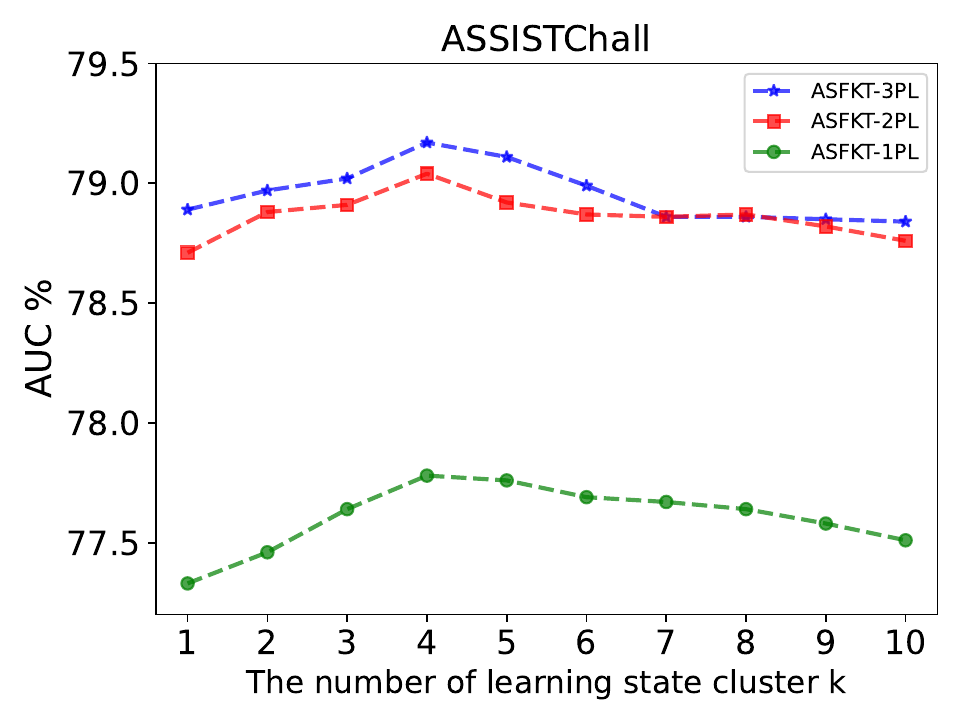}
  \end{minipage}\hspace{-2.2em}
  \begin{minipage}[b]{0.47\textwidth}
    \hspace{-3.2em}
    \includegraphics[width=0.83\linewidth]{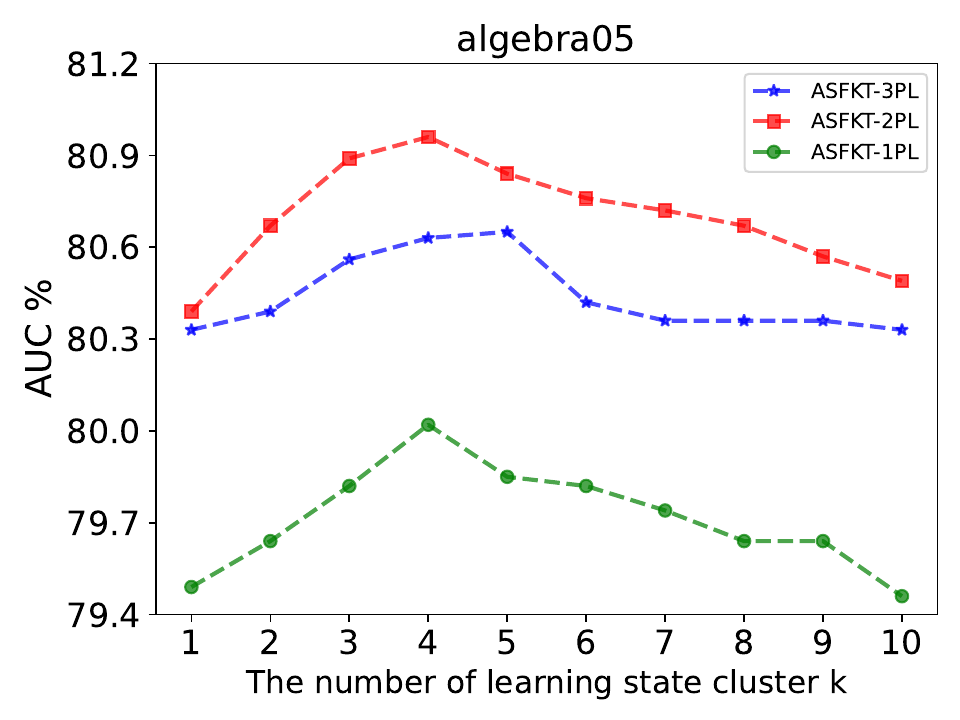}
  \end{minipage}
\caption{
The influence of different clustering quantities on the AUC values of three embedding models. The horizontal axis shows the change in k values from 1 to 10, while the vertical axis displays the percentage of AUC values. Three different colors distinguish the three embedding methods in the graph.}
% 不同聚类数量对三种嵌入模型AUC值的影响。横轴显示k值从1增加到10的变化，纵轴显示AUC值的百分比。图中以三种不同颜色区分了三种不同的嵌入方法。
  \label{FIG:4}
  \vspace{-1em}
\end{figure*}
\par{
To investigate the impact of varying the number of clustering clusters $n$ on the knowledge tracing performance of LSKT, we systematically increased $n$ from 1 to 10 and documented the corresponding changes in AUC across four datasets. The experimental findings, depicted in \Cref{FIG:4}, showcase the relationship between the number of clustering clusters and the AUC percentage. Here, the horizontal axis delineates the progression of clustering clusters, while the vertical axis illustrates the fluctuations in AUC. The three distinctively colored lines denote three diverse embedding modeling methodologies. Upon scrutinizing the outcomes, we draw the following conclusions: irrespective of the embedding modeling approach, an optimal AUC is attained when $n$ is set to 4 across all four datasets. Nevertheless, deviations from this value, either upwards or downwards, lead to a decline in clustering accuracy. Specifically, when $n$ is less than 4, certain inconsequential historical instances are still accommodated by the model via the softmax function, thus introducing superfluous noise during model training. Conversely, when $n$ exceeds 4, pivotal moment information tends to be overlooked. Hence, to ensure the optimal performance of our proposed LSKT, we consistently maintain $n$ at 4 throughout the entirety of this paper's experiments.
% 为了探究聚类簇个数$n$对LSKT知识追踪性能的影响，我们将$n$从1逐步提升到10，并在四个数据集上记录了AUC的变化情况。实验结果如图4所示，其中横轴代表聚类簇数的变化，纵轴代表AUC指标百分比的变化，三种不同颜色的线则分别代表三种不同的嵌入建模方式。通过对结果的观察，我们得出以下结论：无论是哪种嵌入建模方式，当$n$为4时，LSKT在所有四个数据集上的AUC几乎都达到了最高值。而随着$n$的增大或减小，聚类结果的准确性都会下降。具体来说，当$n$小于4时，一些无关的历史时刻仍会通过softmax函数平滑地被模型关注，从而给模型训练带来额外噪声；而当$n$大于4时，一些关键的时刻信息则会被忽视。因此，为了确保我们提出的LSKT能够发挥最佳性能，在全文的实验中，我们始终将$n$设置为4。
}
\subsection{Fine-grained knowledge state (RQ3)}
\par{
To capture more fine-grained knowledge states, this paper firstly models three different types of feature embeddings. However, considering only the differences at the level of problem features and interaction features cannot fully simulate the complex answering process of humans. In the real answering process, the change in the learner's state is also a hidden factor that needs to be considered. For example, even if there are multiple exercises similar to the current one in the historical sequence, the learner's current learning state may be significantly different from the learning state when answering some similar exercises in history due to the continuous change in the process of doing the exercises. Intuitively, if the learner answers a exercise incorrectly, the quality of the learner's state may reflect different levels of mastery of this exercise. Therefore, it is necessary to introduce the change in learning state into the process of extracting knowledge states.
% 为了捕获更细粒度的知识状态，本文首先建模了三种不同的特征嵌入方式，但是仅考虑习题特征与交互特征层面的差异性并不能充分模拟人类复杂的答题过程。在真实的答题过程中，学习者状态的变化也是一个需要被考虑在内的隐藏因素。例如，即使历史序列上存在多道问题与当前需要回答的问题相似，但由于学习者的状态会随着做题过程不断变化，其当前的学习状态可能与历史上在回答某几道相似问题时的学习状态有较大的差异，根据直觉，如果学习者错误回答了某道问题，那么该名学习者状态的好坏可能反映了对这道题不同的掌握程度。所以有必要将学习状态的变化引入知识状态的提取过程中。

To address the issue, a learning state enhanced knowledge state extraction module is proposed in this approach. This module further distinguishes the differences in learners' historical learning states during the process of capturing knowledge states. This module emphasizes that when capturing the learner's knowledge state, it is necessary to not only consider the similarity between the current required answer exercises and historical exercises, but also to consider whether the learner's learning state at the current moment is consistent with that at the corresponding historical moment. By integrating changes in the learning state into the process of capturing learner knowledge states, we can obtain a more detailed understanding of the knowledge state.
% 为了解决上述问题，一种学习状态增强的知识状态提取模块被本方法提出，该模块在获取知识状态的过程中进一步对学习者的历史学习状态的差异进行了区分。该模块强调，在捕获学习者的知识状态时，不仅需要考虑当前所需回答习题与历史习题之间的的相似性关系，还需要考虑学习者在当前时刻与历史对应时刻的学习状态是否一致。通过将学习状态的变化融入到学习者知识状态的捕获过程中，我们能够获得更细粒度的知识状态。
}

\begin{figure*}
  \begin{minipage}[b]{0.256\textwidth}
  \hspace{-1em}
  \centering
    \includegraphics[height=1.05\linewidth,width=1.18\linewidth]{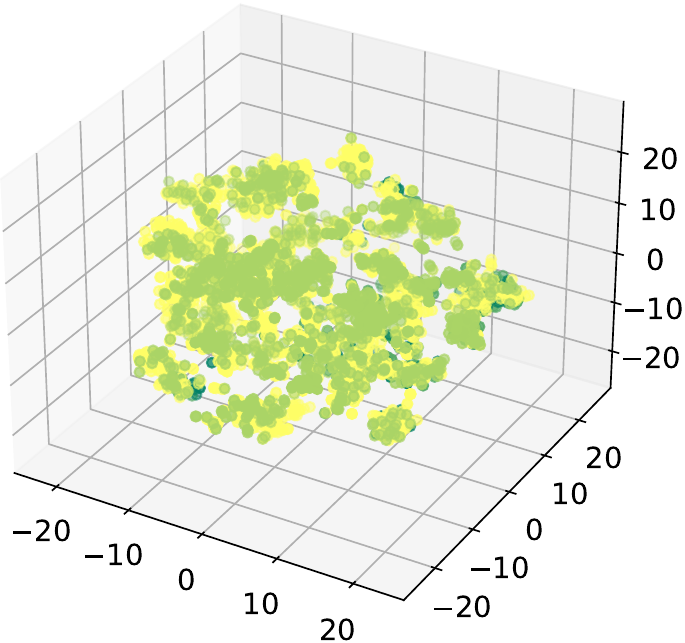}
    (a) Knowledge states\\
    w/o LE
  \end{minipage}\hspace{3.01em}
  \begin{minipage}[b]{0.256\textwidth}
  \hspace{-1em}
  \centering
    \includegraphics[height=1.05\linewidth,width=1.18\linewidth]{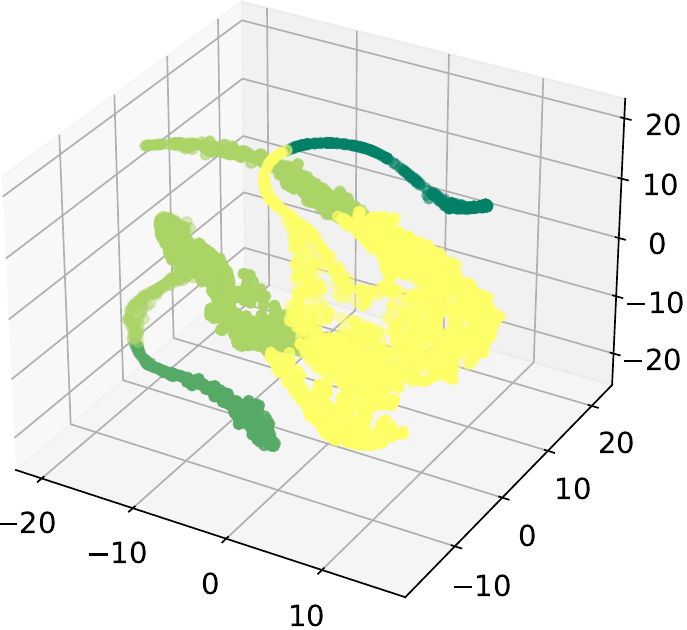}
    (b) Learning states\\ \
  \end{minipage}\hspace{3.01em}
  \begin{minipage}[b]{0.256\textwidth}
  \hspace{-1em}
  \centering
    \includegraphics[height=1.05\linewidth,width=1.18\linewidth]{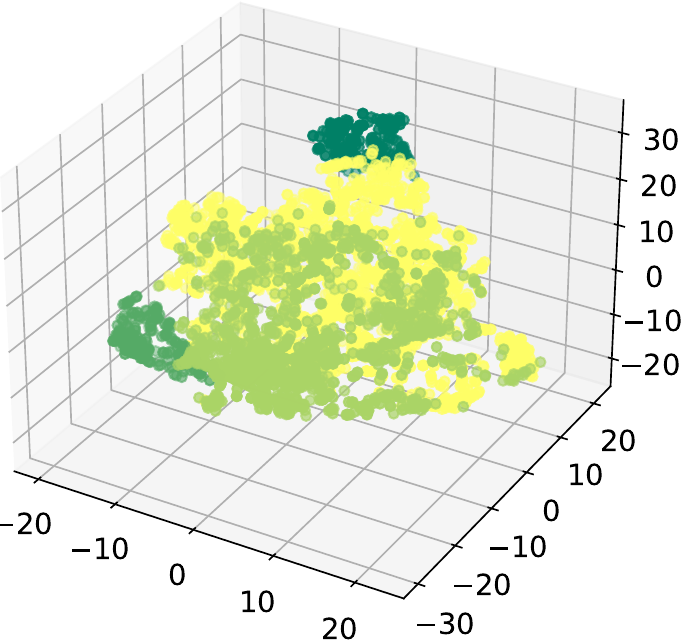}
    (c) Knowledge states\\ \
  \end{minipage}
  \begin{minipage}[b]{0.1\textwidth}
  \hspace{2.3em}
    \includegraphics[height=2.2\linewidth]{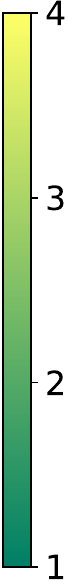}
    \vspace{4em}
  \end{minipage}
\caption{Visualization of Knowledge State and Learning State. Figure (a) represents the distribution of knowledge state features extracted without the guidance of a learning state. Figure (b) illustrates the feature distribution corresponding to the learning state. Figure (c) depicts the distribution of knowledge state features extracted with the guidance of a learning state.}
    % 知识状态与学习状态的可视化。图（a）代表在没有学习状态的引导的情况下提取出的知识状态特征分布情况。图（b）表示对应学习状态的特征分布。图（c）表示在学习状态的引导下提取出的知识状态特征分布。
  \label{FIG:5}
  \vspace{-1em}
\end{figure*}

\par{
In \Cref{FIG:5}, a random selection of 64 learners' knowledge state sequences from the ASSIST09 dataset's test set was made, and T-SNE~\citep{[25]} technology was employed for visualization. Different features of learning state classes were indicated by different colors. \Cref{FIG:5}(a) demonstrates that without considering the learning states, the knowledge state features tend to mix with each other. \Cref{FIG:5}(b) displays the distribution of learning state features at each time step when considering the learning states. On the other hand, \Cref{FIG:5}(c) shows the distribution of knowledge state features after incorporating the learning states into the knowledge state extraction process. It can be observed that by introducing learning states as guidance, the knowledge states are classified more precisely. Specifically, under the same learning state, the learning state features at corresponding time steps tend to cluster together rather than being mixed. This result confirms the importance of learning states in interaction performance. By guiding with learning states, we are able to explore more intrinsic and fine-grained knowledge states.
% 如图5所示，在ASSIST09数据集的测试集中，随机选取了64位学习者的知识状态序列，并使用T-SNE[25]技术对其进行了可视化处理，不同的学习状态类的特征通过不同的颜色进行了标示。图5（a）展示了在未考虑学习状态的情况下，知识状态特征倾向于相互混合。图5（b）则显示了考虑学习状态时，各个时刻的学习状态特征分布情况。而图5（c）展示了在将学习状态纳入知识状态提取过程后，知识状态特征的分布。可以观察到，通过引入学习状态作为指导，知识状态得到了更精细的分类，即在同一学习状态下，对应时刻的学习状态特征更倾向于聚集在一起，而不是相互混淆。这一结果验证了学习状态在交互表现中的重要性，通过学习状态的引导能够挖掘出更为内在和细粒度的知识状态。
}

\subsection{Embeddings visualization(RQ4)}
\par
\begin{figure*}
\begin{minipage}[b]{0.2529\textwidth}
    \hspace{-2em}
    \includegraphics[height=0.89\linewidth]{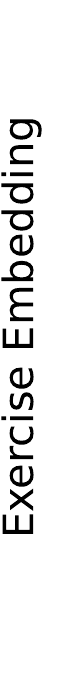}
  \end{minipage}\hspace{-14.5em}
  \begin{minipage}[b]{0.245\textwidth}
    \centering
    \includegraphics[height=0.91\linewidth,width=0.98\linewidth]{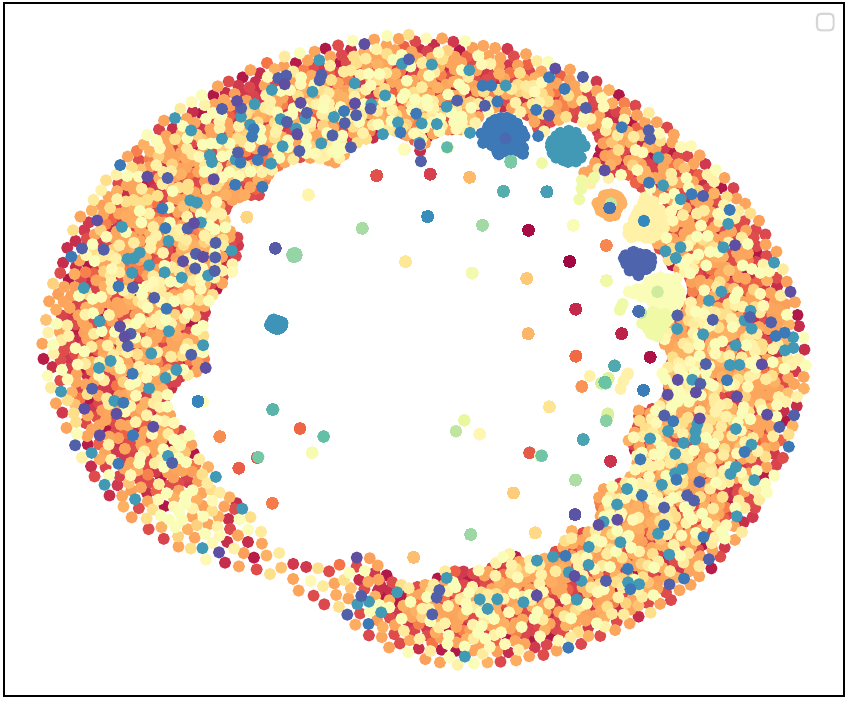}
  \end{minipage}\hspace{-0.2em}
  \begin{minipage}[b]{0.245\textwidth}
    \centering
    \includegraphics[height=0.91\linewidth,width=0.98\linewidth ]{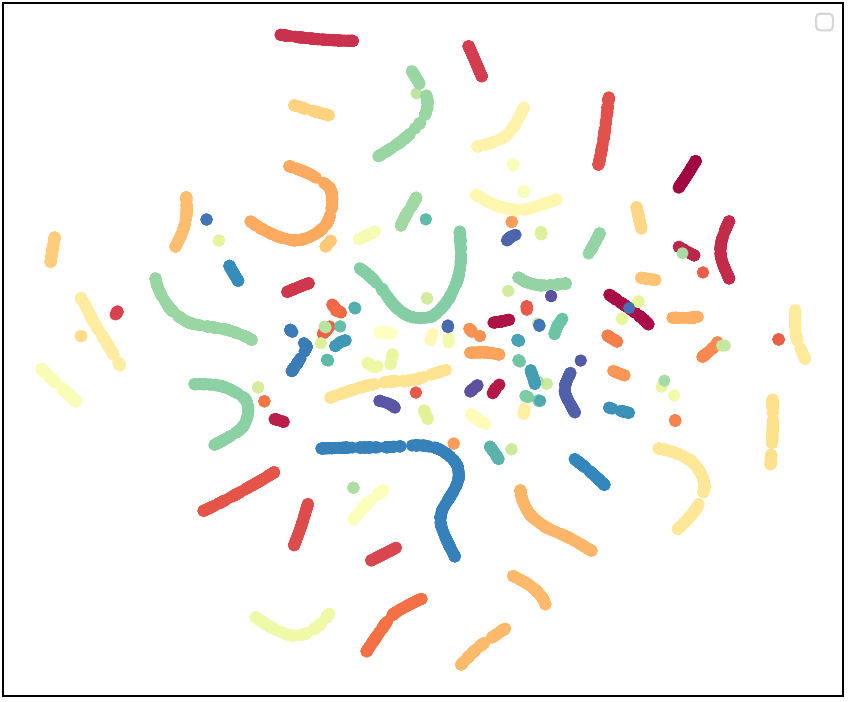}
  \end{minipage}\hspace{-0.2em}
  \begin{minipage}[b]{0.245\textwidth}
    \centering
    \includegraphics[height=0.91\linewidth,width=0.98\linewidth]{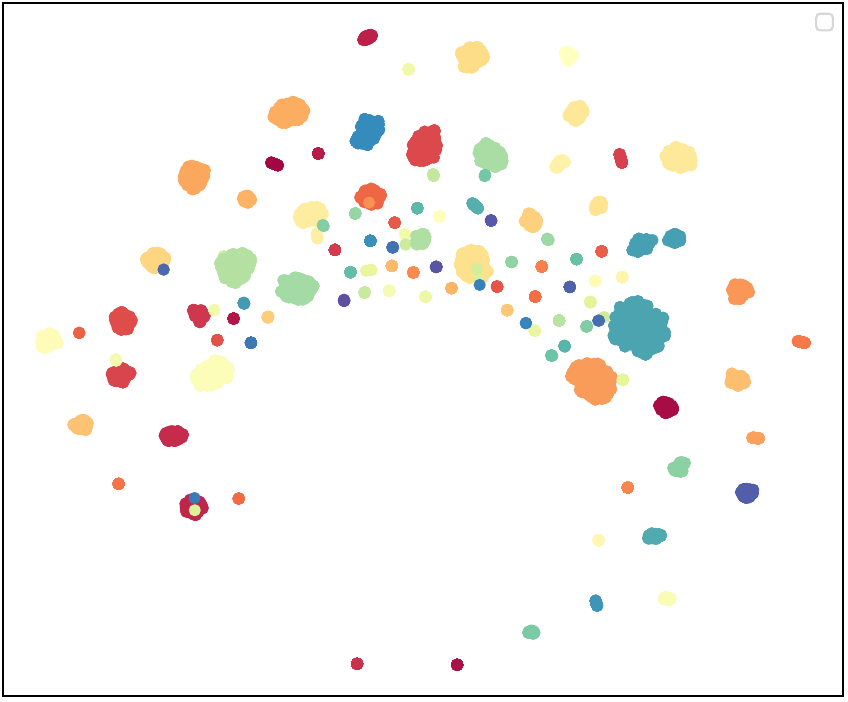}
  \end{minipage}\hspace{-0.1em}
  \begin{minipage}[b]{0.245\textwidth}
    \centering
    \includegraphics[height=0.91\linewidth,width=0.98\linewidth]{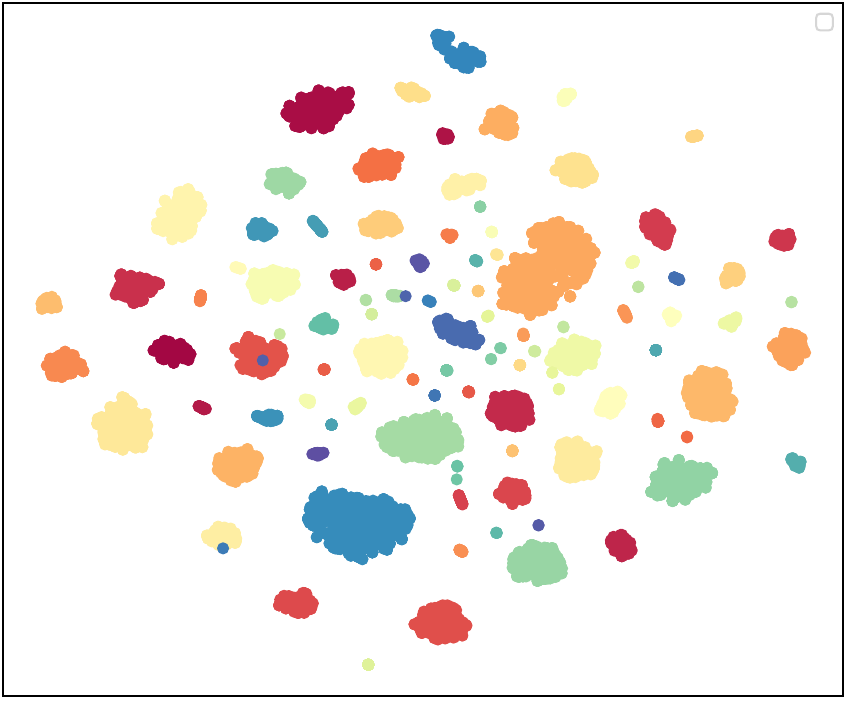}
  \end{minipage}

\vspace{0.4em}
\begin{minipage}[b]{0.2529\textwidth}
\hspace{-2em}
    \includegraphics[height=0.89\linewidth]{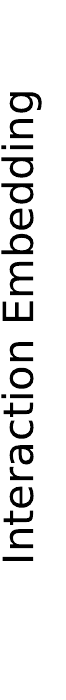}
    \vspace{1em}
  \end{minipage}\hspace{-14.5em}
  \begin{minipage}[b]{0.245\textwidth}
    \centering
    \includegraphics[height=0.91\linewidth,width=0.98\linewidth]{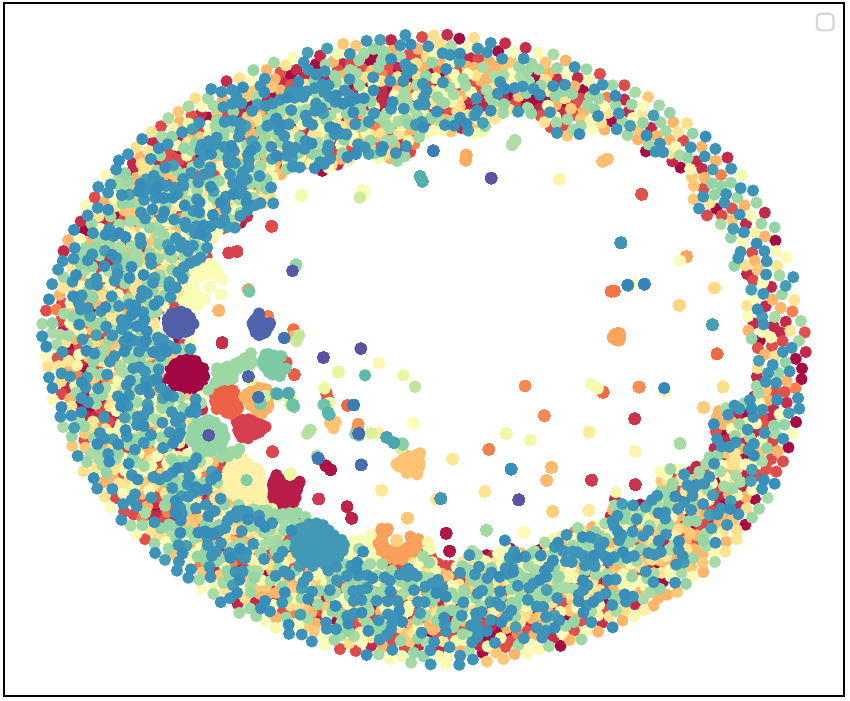}
    (a) LSKT-NI
  \end{minipage}\hspace{-0.2em}
  \begin{minipage}[b]{0.245\textwidth}
    \centering
    \includegraphics[height=0.91\linewidth,width=0.98\linewidth]{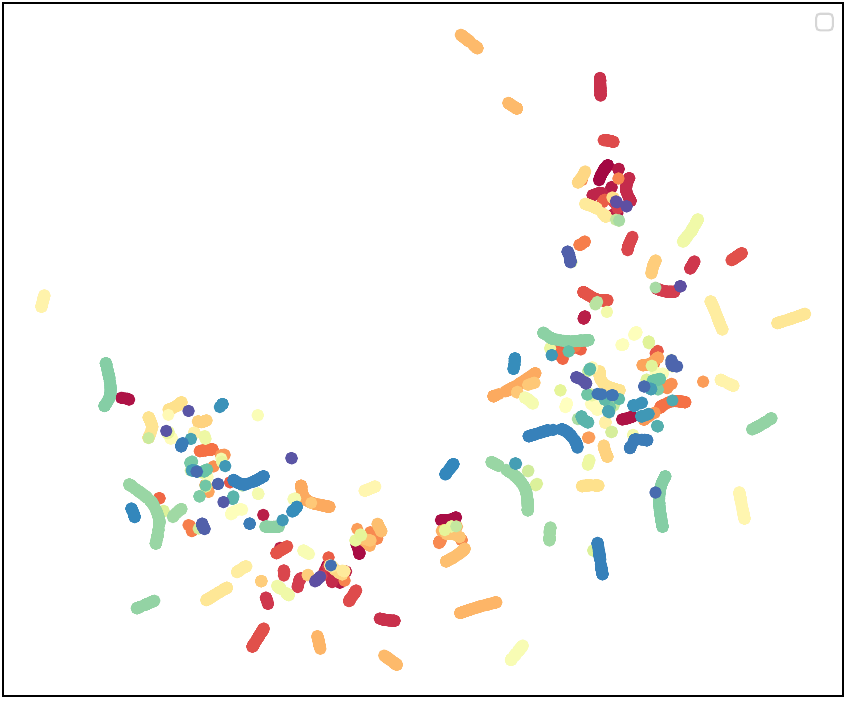}
    (b) LSKT-1PL
  \end{minipage}\hspace{-0.2em}
  \begin{minipage}[b]{0.245\textwidth}
    \centering
    \includegraphics[height=0.91\linewidth,width=0.98\linewidth]{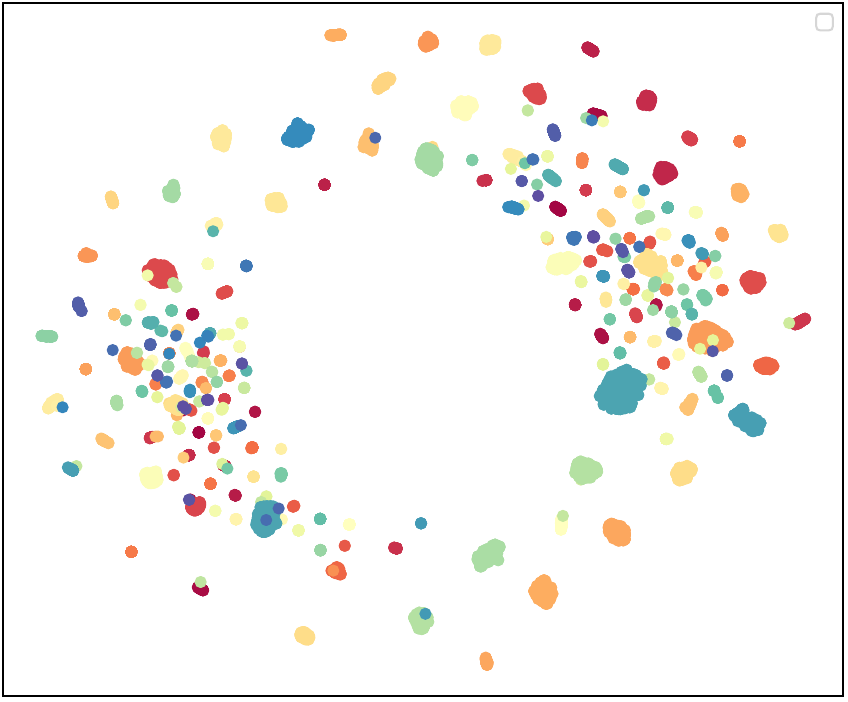}
   (c) LSKT-2PL
  \end{minipage}\hspace{-0.2em}
  \begin{minipage}[b]{0.245\textwidth}
    \centering
    \includegraphics[height=0.91\linewidth,width=0.98\linewidth]{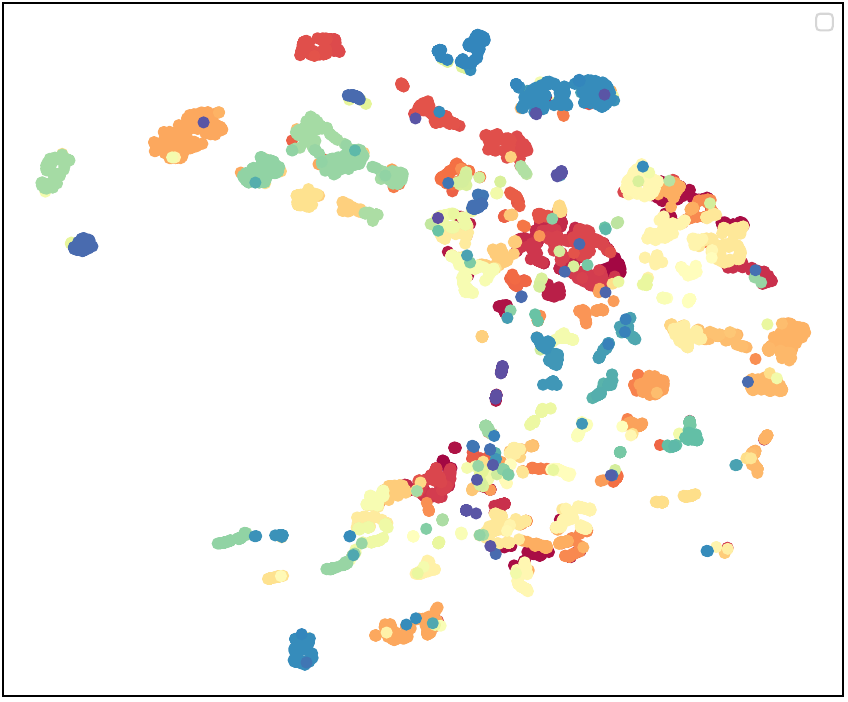}
    (d) LSKT-3PL
  \end{minipage}
\caption{Comparison of four embedding methods on the ASSIST09 dataset. Figure (a) depicts using only conceptual knowledge for embedding features, while figures (b), (c), and (d) illustrate three proposed embedding modeling approaches. The upper and lower graphs respectively show exercise and learner interaction embedding results, with exercises sharing the same knowledge concepts highlighted in the same color.}
% ASSIST09数据集上四种嵌入方式的可视化比较，图（a）表示仅使用知识概念信息作为嵌入特征，图（b）（c）（d）分别对应我们所提出的三种嵌入建模。上方的图与下方图分别展示了习题嵌入特征与学习者交互嵌入特征的可视化结果，拥有相同知识概念的练习用同种颜色标记。
  \label{FIG:6}
  \vspace{-1em}
\end{figure*}

{
We utilize the T-SNE algorithm to perform dimensionality reduction and visualize embeddings containing only exercise concept information, as well as embeddings proposed by LSKT based on modeling differences in three levels of granularity. The aim of this approach was to assess the impact of modeling differences between exercises and between interactions on the interpretability of model embedding features. On the ASSIST09 test set, an equal number of exercises were randomly selected for this visualization experiment, and the results are presented in \Cref{FIG:6}.
% 我们使用T-SNE算法，在二维空间中对仅含有习题概念信息的嵌入以及LSKT提出的基于三种差异度建模的嵌入进行降维可视化分析。这样做是为了评估习题间与交互间差异性建模对模型嵌入特征可解释性的影响。在ASSIST09测试集上，随机选择了等量的习题进行了这一可视化实验，结果如图6所示。
}
\par{
As shown in \Cref{FIG:6}(a), a large number of exercise embedding features and interaction embedding features are mixed together, indicating that if only the concept of exercises is considered without modeling differences, the model will struggle to distinguish certain concept features, thereby increasing the difficulty of capturing effective associative information between exercises.
% 如图6（a）所示，可以看到大量的习题嵌入特征与交互嵌入特征被混淆在一起，这表明如果仅考虑习题的概念而不考虑差异性建模，模型将难以区分某些概念特征，进而增加了模型捕获习题间有效关联信息的难度。

To address the above issues, we attempted to gradually model the differences between interactions. Inspired by the IRT-1PL model, new exercise embedding method and corresponding interaction embedding method were designed. The visualization results, as shown in \Cref{FIG:6}(b), indicate that exercise features and interaction features are linearly clustered in the two-dimensional space. This suggests that the LSKT-1PL method can enable the model to capture some continuity relationships in exercise concepts, such as the progressive relationship of exercise difficulty. However, due to this linear clustering, there is a lack of clear differentiation boundaries between exercise and interaction features. As a result, some exercise representations intertwine with each other in the feature space, making it difficult for the model to determine the relationship and distinction between exercises and interactions.
% 为了解决上述问题，我们尝试逐步建模交互之间的差异性。受IRT-1PL模型的启发，设计了新的习题嵌入方法和与之对应的交互嵌入方法。其可视化结果如图6（b）所示，可以看到习题特征和交互特征在二维空间中呈线状聚集分布，表明LSKT-1PL方法能使模型捕捉到一些习题概念中的连续性关系，例如习题难度的递进关系。然而，这种线状聚集导致习题和交互特征缺乏清晰的区分边界，并使一些习题表征在特征空间中相互交织，使得模型难以确定习题之间与交互之间的联系和区别。

 Inspired by the IRT-2PL model, we designed a new embedding method that further refines the modeling of feature differences. The visualization results, as shown in \Cref{FIG:6}(c), demonstrate that exercises with the same knowledge concepts are clustered closely together in the feature space, while those with different concepts are farther apart. This indicates that the model can effectively differentiate between exercises and interactions with different concepts. However, although differential modeling has been applied to exercises and interactions with the same concepts, the embedding features between exercises and interactions with the same concept still appear very similar, closely aggregated together. This suggests the possibility of overfitting in the model, which is not the expected outcome.
% 受IRT-2PL模型的启发，我们设计了新的嵌入方法，进一步细化了特征的差异性建模。其可视化结果如图6（c）所示，可以看到具有相同知识概念的习题在特征空间中彼此靠近，形成紧密的集群，而包含不同概念的习题相互远离，表明模型能够顺利区分出不同概念的练习与交互。然而，虽然对相同概念的练习之间与交互之间进行了差异性建模，但同概念习题与交互之间的嵌入特征仍然非常相似，紧密的聚合在一起，模型存在过拟合的嫌疑，这并非预期的结果。

Finally, we chose to introduce the learner's guess factor into the interactive sequence embedding. By simulating the unreliability of responses, we reduced the degree of overfitting of the model on unreliable interactions. The visualization results, as shown in \Cref{FIG:6}(d), demonstrate that the clusters formed by exercise features of the same concept are no longer as concentrated as in \Cref{FIG:6}(c), and the differences between different exercises can be better identified. This explains why the LSKT-3PL model performs better on most datasets.
% 最后，我们选择在交互序列嵌入中引入学习者猜测因素。通过模拟responses的不可靠性，减少了模型对不可靠交互的过拟合程度。其可视化结果如图6（d）所示，可以观察到同概念练习特征形成的簇不再像图6（c）中那么集中，不同练习之间的差异性能够更好地被识别出来，这就解释了为什么LSKT-3PL模型在多数数据集上表现更好。

Through visualizing embedding features without modeling differences and respectively modeling differences at coarse-grained, sub-fine-grained, and fine-grained levels,  we can observe the impact of the differential modeling on the ATT-DLKT model. This validates the importance of differentially modeling embedding features for the knowledge tracing task.
% 通过可视化不进行差异性建模的嵌入特征与分别进行粗粒度、次细粒度、细粒度差异性建模的嵌入特征，可以观察到差异性建模对于ATT-DLKT模型的影响，这验证了嵌入特征的差异性建模对于知识追踪任务的重要性。
}

\subsection{Visualization of knowledge state variation }
\par
\begin{figure*}
\begin{minipage}[b]{1\textwidth}
\hspace{4.85em}
    \includegraphics[width=0.826\linewidth]{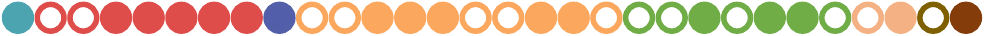}
  \end{minipage}

  \begin{minipage}[b]{0.15\textwidth}
  \hspace{-1.3em}
    \includegraphics[height=0.9\linewidth]{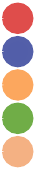}
  \vspace{0.3em}
  \end{minipage}\hspace{-7.9em}
  \begin{minipage}[b]{0.9\textwidth}
  \hspace{-2.7em}
    \centering
    \includegraphics[height=0.163\linewidth,width=0.94\linewidth]{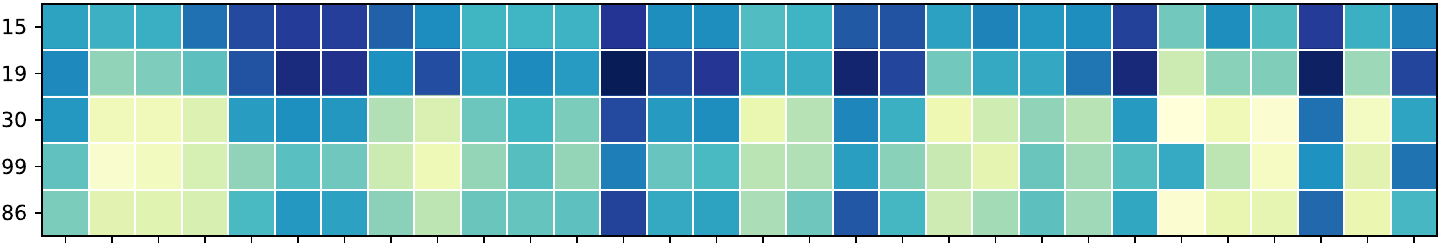}
  \end{minipage}\hspace{-0.8em}
  \begin{minipage}[c]{0.02\textwidth}
    \centering
  \vspace{-7em}
    \includegraphics[height=10.95\linewidth,width=1.8\linewidth]{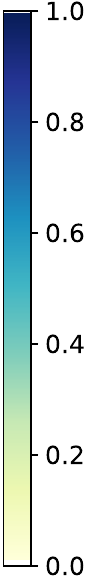}
  \end{minipage}\hspace{-0.47em}
  \begin{minipage}[c]{0.02\textwidth}
  \vspace{-7em}
    \centering
    \includegraphics[height=11\linewidth,width=2\linewidth]{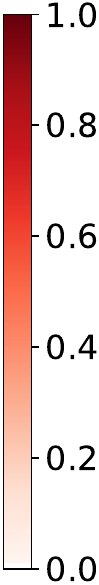}
  \end{minipage}

  \vspace{-1.9em}
  \begin{minipage}[b]{0.9\textwidth}
\centering
  \hspace{-4.5em}
    \includegraphics[height=0.055\linewidth,width=0.915\linewidth]{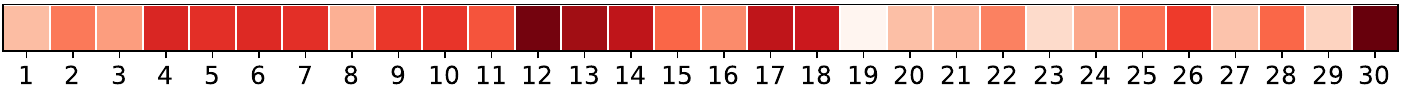}
    
(a) The change process of learner knowledge state (blue) to learning state (red)
  \end{minipage}

\vspace{1.6em}
\begin{minipage}[b]{1\textwidth}
\hspace{4.6em}
    \includegraphics[width=0.84\linewidth]{vispro_up.pdf}
  \end{minipage}

\begin{minipage}[b]{0.15\textwidth}
  \hspace{-1.6em}
    \includegraphics[height=0.75\linewidth]{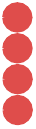}
   \vspace{3.6em}
  \end{minipage}\hspace{-6.7em}
  \begin{minipage}[b]{0.9\textwidth}
  \hspace{-3.4em}
    \centering
    \includegraphics[height=0.15\linewidth,width=0.98\linewidth]{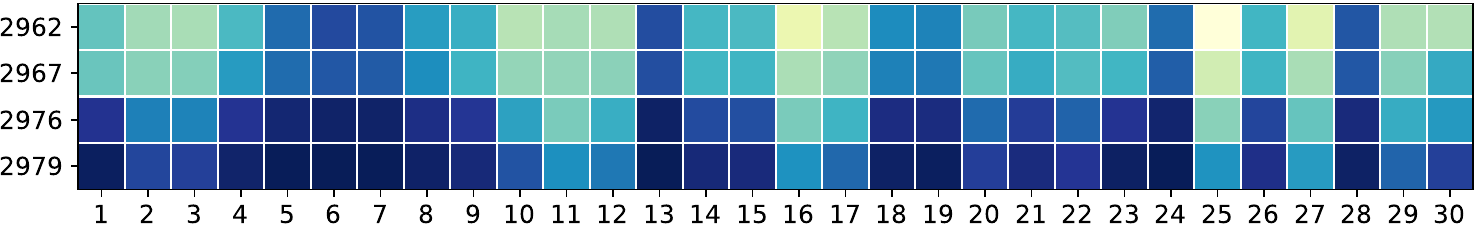}\\
      (b) Knowledge state changes on the same learning sequence for

four different exercises containing 15
  \end{minipage}\hspace{-0.1em}
  \begin{minipage}[c]{0.02\textwidth}
  \vspace{-13.9em}
    \centering
    \includegraphics[height=9.5\linewidth,width=1.9\linewidth]{blue_bar.pdf}
  \end{minipage}
\caption{The different knowledge concepts in the graph correspond to circles of different colors. The circle at the top of the image represents the knowledge concept of the exercise being done at that moment. Hollow circles indicate that the learner answered incorrectly, while solid circles indicate that the learner answered correctly. Figure (a) shows the evolution of the learner's knowledge state (blue) and learning state (red) during the continuous 30 exercises, where the deeper the color, the better the learner's mastery of the knowledge concept or the better the learning state. Figure (b) shows the differences in the evolution of mastery level on 5 different exercises involving knowledge concept 15, where each row number represents the real exercise number.}
% 图中不同知识概念对应不同颜色的圆圈，图片上方的圆圈表示该时刻所做习题的知识概念。空心圆圈表示学习者回答错误，实心圆圈表示学习者回答正确。图(a)中展示了学习者在做连续的30道练习时的知识状态（蓝色）和学习状态（红色）的演化过程，其中颜色越深代表学习者对该知识概念的掌握越好或学习状态越好。图（b）展示了在包含知识概念15上的的5道不同练习上的习题掌握程度演化的差异，其中每行的标号代表的是真实题号。
  \label{FIG:7}
\end{figure*}

{
Predicting the changes in knowledge state is one of the important goals of knowledge tracing tasks. To explore the effect of exercise difference modeling and learning state on the evolution of knowledge state, we show a visualization example of the changes in knowledge state on the continuous 30-exercise records of 5 knowledge concepts in the ASSIST09 dataset in \Cref{FIG:7}. The indexes of these 5 knowledge concepts are $\{$15, 19, 30, 99, 86$\}$ respectively. Each row in the figure shows the estimated evolution of the knowledge state of a single corresponding knowledge concept in the answer sequence. Each column shows the changes in knowledge and learning state after completing one exercise answer.
% 预测知识状态的变化是知识跟踪任务的重要目标之一。为了探究习题差异性建模和学习状态对知识状态演变的影响，我们在图7中展示了ASSIST09数据集中5个知识概念的连续30个答题记录上的知识状态变化的可视化实例。这5个知识概念的索引分别为{15，19，30，99，86}，图中每一行显示了单个对应知识概念在答题序列中的知识状态演化估计，每一列表示完成一次习题回答后的知识状态与学习状态的变化。
}
\par{
\Cref{FIG:7}(a) illustrates the evolution of learners' knowledge and learning states while completing a continuous sequence of 30 exercises. The blue process at the top represents the evolution of learners' mastery levels of individual skills. It can be observed that when learners correctly (incorrectly) complete an exercise, the mastery level of the corresponding knowledge concept increases (decreases). The red process at the bottom represents the evolution of learners' learning states. It can be seen that learning states change as the learning process progresses; when learners frequently answer exercises correctly, their learning state improves, and when they frequently answer exercises incorrectly, their learning state declines, which aligns with common intuition. Our LSKT simultaneously considers the influence of both knowledge states and learning states on learners' performance, which better reflects the real answering process.
% 图7（a）中展示了学习者在做连续的30道习题时知识状态与学习状态的演化过程。上方的蓝色的变化过程代表学习者对单个技能掌握程度的演变。可以看到，当学习者正确（错误）地完成一个练习时，相应知识概念的掌握水平会增加（减少）；下方红色的变化过程代表学习者的学习状态的演变。可以看到，学习状态会随着学习过程的进行而变化，当学习者频繁答对题目时，其学习状态会有所提升，当学习者频繁答错题目时，其学习状态也会下降，这符合大众的普遍直觉。我们的LSKT同时考虑了知识状态与学习状态对学习者的答题表现的影响，这更符合真实的答题过程。
}
\par{
In \Cref{FIG:7}(b), we show the differences in the evolution of knowledge mastery for different exercises with the same knowledge concept $\{$15$\}$, which are $\{$2962, 2967, 2976, 2979$\}$. Although these exercises belong to the same knowledge concept, the evolution of their knowledge mastery is not exactly the same. This difference more closely aligns with our perception that, even though exercises belong to the same concept, their mastery levels can differ due to the complexity of different factors such as difficulty and discrimination of the exercises. This variability in modeling needs to be considered.
% 图7（b）中展示了具有相同知识概念（15）的不同习题{2962，2967，2976，2979}间知识掌握程度的演变差异。尽管这些习题都属于同一知识概念，但它们的知识掌握程度的演变并不完全相同。这种差异也符合我们的直觉，即尽管习题属于同一概念，但由于不同习题的难度、辨识度等复杂因素的差异，它们的知识掌握程度也会有所不同。这种差异性建模需要被考虑在内。
}

\par{
Through this experiment, we can see that LSKT is able to assess learners' mastery of each knowledge concept effectively. By modeling the differences between exercises and learners' actual learning states, LSKT can provide a more nuanced reflection of the genuine process of learners' knowledge acquisition.
% 通过这个实验，我们可以看到LSKT能够良好评估学习者对每个知识概念的掌握程度。通过对习题之间的差异和学习者真实的学习状态的建模，LSKT能够更精细地反映学习者的真实知识掌握过程。
}
\subsection{Discussion and conclusion}
\par{
This paper explores two crucial factors that are often overlooked in the existing ATT-DLKT model, namely, the modeling of latent differences in interactive features and the changes in learners' state during the answering process. Specifically, we propose a more fine-grained model, LSKT, which models these two factors during the feature extraction and training process, and integrates them into the KT task. This not only improves the accuracy of the model, but also models a process more in line with the actual answering process. In addition, this paper conducts a visual comparative study of knowledge concept embedding features and three different granularities of exercise and interaction embedding features, the results of which highlight the importance of differential modeling in the feature extraction process. Experimental results on four standard datasets prove that our model outperforms other baseline models.
% 本文探讨了现有的ATT-DLKT模型经常忽略但却很关键的两个因素，即交互特征的潜在差异性建模与答题过程中学习者状态的变化。具体而言，本文提出了一种更加细粒度的模型LSKT，该模型在特征提取过程中与训练过程中建模了这两个因素，并将其融入到KT任务中，从而在提高了模型的准确性的同时建模了更符合真实答题过程的模型。此外，本文对知识概念嵌入特征与三种不同粒度的习题与交互嵌入特征进行了可视化对比研究，结果显示了在特征提取过程中进行差异性建模的重要性。在四个标准数据集上的实验结果证明，我们的模型优于其他的基线模型。
}

\par{In our future work, we plan to delve into the differences between exercises under the same knowledge concept and the connections between exercises under different knowledge concepts. We hope to fully exploit the potential information in the dataset to further enhance the performance and applicability of our model.
% 在未来的工作中，我们计划深入探索在相同知识概念下习题之间的差异性，以及在不同知识概念下习题之间的联系。我们希望充分挖掘数据集中存在的潜在信息，从而进一步提升模型的性能和实用性。
}

	\section*{CRediT authorship contribution statement}
	\par{\textbf{Shanshan Wang:} Conceptualization, Methodology, Writing – review \& editing. \textbf{Xueying Zhang:} Methodology, Validation, Investigation, Writing – original draft. \textbf{Xun Yang:} Conceptualization, Supervision, Resources. \textbf{Xingyi Zhang:} Supervision, Resources. \textbf{Keyang Wang:} Supervision, Resources. }

	\section*{Declaration of competing interest}
	\par{The authors declare that they have no known competing financial interests or personal relationships that could have appeared to influence the work reported in this paper.}

	% \section*{Funding}
	% \par{This work is supported by National Natural Science Fund of China (No. 62106003, 62272435) and Joint Funds of the National Natural Science Foundation of China (U22A2094).}

	% 致谢
	\section*{Acknowledgements}
	\par{This work is supported by National Natural Science Fund of China (No. 62106003, 62272435) and Joint Funds of the National Natural Science Foundation of China (U22A2094).}

	\bibliographystyle{model5-names}
	% \bibliographystyle{cas-model2-names}

	% \bibliography{LSKT}


	%Author biography without author photo.


\end{sloppypar}
\end{document}